\documentclass[10pt,twocolumn,letterpaper]{article}

\usepackage[pagenumbers]{cvpr}              % To produce the CAMERA-READY version
\newcommand\blfootnote[1]{%
  \begingroup
  \renewcommand\thefootnote{}\footnote{#1}%
  \addtocounter{footnote}{-1}%
  \endgroup
}

\usepackage[dvipsnames]{xcolor}
\usepackage{amssymb}

\definecolor{myblue}{RGB}{68, 114, 196}

\definecolor{pipgreen}{RGB}{112, 173, 71}
\definecolor{pipblue}{RGB}{0, 176, 240}
\usepackage{amsmath,amsfonts,bm}

 \def\eqref#1{(\ref{#1})}
\def\1{\bm{1}}

\def\rvepsilon{{\mathbf{\epsilon}}}
\def\rvtheta{{\mathbf{\theta}}}

\def\vzero{{\bm{0}}}

\def\vmu{{\bm{\mu}}}

\def\vx{{\bm{x}}}
\def\vy{{\bm{y}}}
\def\vz{{\bm{z}}}

\def\mI{{\bm{I}}}

\DeclareMathAlphabet{\mathsfit}{\encodingdefault}{\sfdefault}{m}{sl}
\SetMathAlphabet{\mathsfit}{bold}{\encodingdefault}{\sfdefault}{bx}{n}

\def\gL{{\mathcal{L}}}

\def\gN{{\mathcal{N}}}

\def\sU{{\mathbb{U}}}

\newcommand{\E}{\mathbb{E}}

\usepackage{algorithm}
\usepackage{algorithmic}
\usepackage{multirow}
\usepackage{graphicx}

\definecolor{cvprblue}{rgb}{0.21,0.49,0.74}
\usepackage[pagebackref,breaklinks,colorlinks,citecolor=cvprblue]{hyperref}

\def\paperID{9125} % *** Enter the Paper ID here
\def\confName{CVPR}
\def\confYear{2024}

\usepackage{colortbl}
\title{DREAM: Diffusion Rectification and Estimation-Adaptive Models}

\author{Jinxin Zhou$^{1*}$
\and Tianyu Ding$^{2*\dagger}$
\and Tianyi Chen$^{2}$
\and Jiachen Jiang$^{1}$
\and Ilya Zharkov$^{2}$
\and Zhihui Zhu$^{1}$
\and Luming Liang$^{2\dagger}$
}

\date{{\large $^1$Ohio State University \qquad  $^2$Microsoft}\\
{\tt\small \{zhou.3820,jiang.2880,zhu.3440\}@osu.edu},
{\tt\small \{tianyuding,tiachen,zharkov,lulian\}@microsoft.com}
}
\begin{document}
\twocolumn
[{%
\renewcommand\twocolumn[1][]{#1}%
\maketitle
%\ificcvfinal\thispagestyle{empty}%\fi
\begin{center}
    \vspace{-.25in}
    \includegraphics[width=0.99\linewidth]{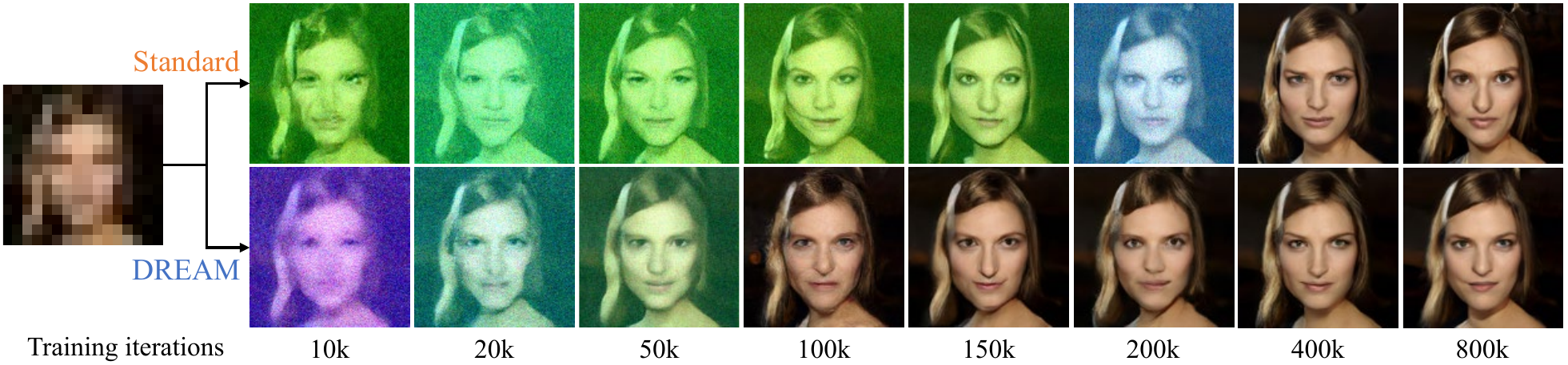}
    {\small Turning \textcolor{orange}{the top} to \textcolor{myblue}{the bottom} by adding only \textbf{\emph{three}} lines of code (line~\ref{line:start}-\ref{line:end} in~\cref{alg:dr-training}).}
    \vspace{-.1in}
    \captionof{figure}{Comparative training of conditional diffusion models for super-resolution. Top: standard conditional DDPM~\cite{saharia2022image}. Bottom: enhancing the same model training with just \emph{three} additional lines of code, leaving the sampling process unchanged. DREAM facilitates notably faster and more stable training convergence, significantly surpassing baseline models in key metrics of perception and distortion.}\label{fig:teaser}
  \end{center}
}]

\begin{abstract}
% Diffusion-based image super-resolution (SR) has gained significant attention for its generative quality, yet a notable training-sampling discrepancy in standard diffusion models limits their full potential. We introduce the Diffusion Rectification and Estimation-Adaptive Model (DREAM), a novel training paradigm requiring minimal code changes (just three lines) but offering substantial improvements in aligning training with sampling. DREAM comprises two components: \emph{diffusion rectification}, which adjusts training to reflect the sampling process,  and \emph{estimation adaptation}, which balances perception against distortion. Our experiments show that DREAM significantly outperforms baseline methods across various diffusion-based SR approaches, accelerates training convergence by $2$ to $3\times $,  and enhances sampling efficiency, needing $10$ to $20\times$ fewer steps for comparable or superior results. We hope DREAM will inspire a rethinking of diffusion model training paradigms.

\blfootnote{$^*$Equal contribution. This work was done when Jinxin Zhou was an intern at Applied Sciences Group, Microsoft.}
\blfootnote{$^\dagger$Corresponding author.}\hspace{-.06in}We present DREAM, a novel training framework representing \textbf{D}iffusion \textbf{R}ectification and \textbf{E}stimation-\textbf{A}daptive \textbf{M}odels, requiring minimal code changes (just three lines) yet significantly enhancing the alignment of training with sampling in diffusion models. DREAM features two components: \emph{diffusion rectification}, which adjusts training to reflect the sampling process,  and \emph{estimation adaptation}, which balances perception against distortion. When applied to image super-resolution (SR), DREAM adeptly navigates the tradeoff between minimizing distortion and preserving high image quality.  Experiments demonstrate DREAM's superiority over standard diffusion-based SR methods, showing a $2$ to $3\times $ faster training convergence and a $10$ to $20\times$ reduction in sampling steps  to achieve comparable results. We hope DREAM will inspire a rethinking of diffusion model training paradigms. Our source code is available at \href{https://github.com/jinxinzhou/DREAM}{link}.

\end{abstract}    
\section{Introduction}
\label{sec:intro}

Single-image super-resolution~(SISR)~\cite{sun2010gradient,dong2014learning, bevilacqua2012low, yan2015single} involves generating high-resolution (HR) images from low-resolution (LR) counterparts, a process crucial in various applications including  video surveillance, medical diagnosis, and  photography. SISR is challenging due to the diverse real-world degradation patterns and the inherent ill-posed nature of the task, where different HR images can correspond to the same LR image.
% The single-image super-resolution (SISR) \cite{sun2010gradient,dong2014learning, bevilacqua2012low, yan2015single} entails creating high-resolution (HR) images from given low-resolution (LR) counterparts, which plays an essential role in various applications ranging from video surveillance to medical diagnosis and photography. However, it is a challenging problem due to the unknown various degradation in the real-world and the inherent ill-posed property that different HR images can share the same downsample LR image. Over the past decade, deep learning algorithms have significantly enhanced state-of-the-art in this area, with progress primarily divided into two categories: regression-based and generation-based methods.
% \jz{SR; difficulties and progression.}

SISR methods are generally categorized into regression-based and generation-based approaches.   Regression-based methods~\cite{lim2017enhanced,zhang2018residual,liang2021swinir,chen2021learning}  focus on minimizing pixel-level discrepancies, \ie, distortion, between SR predictions and HR references. However, this approach often fails to capture the perceptual quality of images. To address this, generation-based methods employ deep generative models,  including autoregressive models~\cite{oord-arxiv-2016,oord-nips-2016}, variational autoencoders~(VAEs)~\cite{Kingma2013,vahdat2021nvae}, normalizing flows~(NFs)~\cite{dinh2016density,Kingma2018}, and generative adversarial networks (GANs)~\cite{goodfellow2014generative,karras2018ProGAN,radford2015unsupervised, liang2022details},  aiming to improve the perceptual aspects of SR images.

Recently, Diffusion Probabilistic Models (DPMs)~\cite{ho2020denoising,sohl2015deep}, a novel class of generative models, have attracted increased interest for their impressive generative abilities, especially in the SISR task~\cite{saharia2022image,yue2023resshift,gao2023implicit,rombach2022high,ho2022cascaded}. Nonetheless, DPM-based methods face challenges due to their dependence on a long sampling chain, which can lead to error accumulation and reduce training and sampling efficiency. A further issue is the discrepancy between training and sampling~\cite{ning2023input,yu2023debias}: training typically involves denoising noisy images conditioned on ground truth samples, whereas testing (or sampling) conditions on previously self-generated results.  This disparity, inherent in the multi-step sampling process, tends to magnify with each step, thereby constraining the full potential of DPMs in practice. 
% \zz{I like the current introduction. Do we want to add one sentence about training and sampling efficiency issues, since we highlight DREAM can improve the efficiency several times.}
% Recently, Diffusion Probabilistic Model (DPMs), an novel class of generative frameworks, has attracted growing interest due to its impressive generation ability and have demonstrated great potential in the SISR task. Unlike proceeding generative methods which creates SR images in a single step, the conditional DDPM model involves a Markov chain, encompassing a forward process that traverses the chain, adding noise to the ground-truth HR images, and a reverse process, which conducts reverse sampling from the chain for denosing from pure Gaussain noise to the HR image, conditioned on the LR image. Analogous to the unconditional case, the presence of estimation error induces a discrepancy between the training and sampling phases. Training is learned towards denoising from noisy images derived from ground-truth data, whereas sampling involves denoising from noisy images generated in prior steps. Additionally, the multi-step nature of sampling means that such discrepancies accumulate with each step, thus curtailing the full potential of DPMs. 

To bridge the gap between training and sampling in diffusion models, we introduce DREAM, an end-to-end training framework denoting Diffusion Rectification and Estimation-Adaptive Models. DREAM consists of two key elements: \emph{diffusion rectification} and \emph{estimation adaptation}. Diffusion rectification extends traditional diffusion training with an extra forward pass, enabling the model to utilize its own predictions. This approach accounts for the discrepancy between training (using ground-truth data) and sampling (using model-generated estimates). However, solely relying on this self-alignment can compromise perceptual quality for the sake of reducing distortion. To counter this, our estimation adaptation strategy balances standard diffusion and diffusion rectification by adaptively incorporating ground-truth information.  This approach smoothly transitions focus between the two by adaptively injecting ground-truth information. This integration harmonizes the advantages of both approaches, effectively reducing the training-sampling discrepancy, as demonstrated in~\Cref{fig:error-dynamic}.
% To align the training process with sampling process, this paper introduces a DREAM framework, a simple but effective end-to-end training strategy for SR task. This framework comprises two fundamental components: \emph{diffusion rectification} and \emph{estimation adaptation}. \emph{Diffusion rectification} extends traditional diffusion training by integrating a single additional forward pass, which allows the model to use its own predictions. This elegant modification empowers denoiser networks to account for the training-sampling discrepancy that results from different constructions of intermediate signals---from ground-truth data during training versus from model-generated estimates during sampling. While diffusion rectification introduce additional supervision to account for the sampling process, directly applying it to the SR task could result in a trade-off between improved distortion and reduced perceptual quality. To harness the strengths of both standard diffusion and diffusion rectification, we propose an \emph{estimation adaptation} strategy that smoothly shifts focus between standard diffusion and diffusion rectification based on the trend of estimation error. 
% This strategy allows the model to integrate the superior quality of standard diffusion with the minimized distortion of diffusion rectification.

The DREAM framework excels in its simplicity, easily integrating into existing diffusion-based models with only three lines of code and requiring no alterations to the network architecture or sampling process. When applied to the SR task, DREAM has notably improved generation quality across various diffusion-based SR methods and datasets. For example, on the $8\times$ CeleA-HQ dataset, it boosts the SR3~\cite{saharia2022image} method's PSNR from $23.85$ dB to $24.63$~dB while reducing the FID score from $61.98$ to $56.01$. Additionally, DREAM accelerates training convergence by $2$ to $3$ times and improves sampling efficiency, requiring $10$ to $20$ times fewer steps for comparable or superior results. It also demonstrates enhanced out-of-distribution (OOD) SR results compared to baseline methods.
% The DREAM framework stands out for its simplicity, allowing for effortless replication and incorporation into existing DPM frameworks with just three lines of code, all without the need for any changes to the network architecture or loss function. Despite its simplicity, as illustrated in \Cref{fig:error-dynamic}, it can effectively alleviate the training-sampling discrepancy. Thanks to the better alignment, our experiments demonstrate that DREAM markedly reliably  enhances the generation quality with a significant margin when employed across a variety of diffusion-based SR methods on different datasets. For instance, on the $4\times$ CeleA-HQ dataset, it improves the baseline PSNR from $23.14$ dB to $24.63$ dB and the FID from $72.23$ to $56.01$. Furthermore, DREAM consistently speeds up training convergence and improves sampling efficiency. 

Our contributions are summarized as follows:
\begin{itemize}
    % \textbf{Simple and generic framework.}
    \item We introduce DREAM, a simple yet effective framework to alleviate the training-sampling discrepancy in standard diffusion models, requiring minimal code modifications.
    \item We demonstrate the application of DREAM to various diffusion-based SR methods, resulting in significant improvements in distortion and perception metrics.
    % \textbf{Enhanced generation quality.}
    \item The proposed DREAM also notably speeds up training convergence, enhances sampling efficiency, and delivers superior out-of-distribution (OOD) results.
    %We establish that DREAM notably speeds up training convergence and enhances sampling efficiency, and it also delivers superior out-of-distribution (OOD) results.
    % \textbf{Faster convergence \& sampling efficiency.}
\end{itemize}

\section{Related work}
\label{sec:related}
\noindent\textbf{Super-resolution.} 
In single-image super-resolution, substantial efforts~\cite{liang2022details, jo2021tackling, ledig2017photo, zhang2018image, soh2019natural,
ding2022sparsity, ding2021cdfi, geng2022rstt,
 zhang2020deep, zhang2021designing, zhang2018residual, anwar2020densely} have been devoted to two primary categories: regression-based and generation-based. Regression-based methods, such as EDSR~\cite{lim2017enhanced}, RRDB~\cite{wang2018esrgan}, and SWinIR~\cite{liang2021swinir}, focus on a direct mapping from LR to HR images, employing pixel-wise loss to minimize differences between SR images and their HR references. While effective in reducing distortion, these methods often yield overly smooth, blurry images. Generation-based methods, on the other hand, aim to produce more realistic SR images. GAN-based models, like SRGAN~\cite{ledig2017photo}, combine adversarial and perceptual losses~\cite{zhang2018unreasonable} to enhance visual quality. Methods of this line include SFTGAN~\cite{wang2018recovering} and GLEAN~\cite{chan2021glean}, which integrate semantic information to improve texture realism. ESRGAN~\cite{wang2018esrgan} further refines SRGAN’s architecture and loss function. However, GAN-based methods often face challenges like complex regularization and optimization to avoid instability. Autoregressive models (\eg, Pixel-CNN~\cite{van2016pixel}, Pixel-RNN~\cite{oord-nips-2016}, VQVAE~\cite{van2017neural}, and LAR-SR~\cite{guo2022lar}) are computationally intensive and less practical for HR image generation. Normalizing Flows (NFs)~\cite{dinh2016density,Kingma2018} and VAEs~\cite{Kingma2013,vahdat2021nvae} also contribute to the field, but these methods sometimes struggle to produce satisfactory results.

\textbf{Diffusion model.}
Inspired by non-equilibrium statistical physics, \cite{sohl2015deep} first proposes Diffusion Probabilistic Models (DPMs) to learn complex distributions. These models have since advanced significantly~\cite{ho2020denoising, song2021denoising, nichol2021improved, dhariwal2021diffusion}, achieving state-of-the-art results in image synthesis.  Beyond general image generation, diffusion models have shown remarkable utility in low-level vision tasks, particularly in SR. Notable examples include SR3~\cite{saharia2022image}, which excels in image super-resolution through iterative refinement, and IDM~\cite{gao2023implicit}, which blends DPMs with explicit image representations to enable flexible generation across various resolutions. SRDiff~\cite{li2022srdiff} uniquely focuses on learning the residual distribution between HR and LR images through diffusion processes. LDM~\cite{rombach2022high} deviates from traditional pixel space approaches, employing cross-attention conditioning for diffusion in latent space. Building upon LDM, ResShift~\cite{yue2023resshift} employs a refined transition kernel for sequentially transitioning the residual from LR embeddings to their HR counterparts. 

\textbf{Training-sampling discrepancy.} 
% Orthogonal to the line of works \cite{song2021denoising, lu2022dpm} of improving sampling efficiency, 
\cite{ning2023input} first analyzes the training-sampling discrepancy in unconditional diffusion models, proposing to represent estimation errors with a Gaussian distribution for improved DPM training.  This discrepancy was later attributed by~\cite{yu2023debias} to a constant training weight strategy, suggesting a reweighted objective function based on the signal-to-noise ratio at different diffusion steps.  In addition, \cite{li2023alleviating} adjusts the distribution during the sampling process by choosing the optimal step within a predefined windows for denoising at each stage. \cite{ning2023elucidating} applies a predefined linear function to adjust noise variance during sampling, and~\cite{everaert2023exploiting} recommends starting the sampling from an approximate distribution that mirrors the training process in terms of frequency and pixel space.

Our approach, distinct from previous unconditional methods, addresses discrepancies based on predictions relative to the conditional input data, ensuring a tailored and accurate solution for complex visual prediction tasks like SISR. Our method also draws inspiration from step-unrolling techniques in depth estimation~\cite{saxena2023monocular, ji2023ddp} and text generation~\cite{savinov2022stepunrolled}, leveraging the model's own predictions for error estimation. However, we uniquely integrate self-estimation with adaptive incorporation of ground-truth data. This integration, guided by the pattern of estimation errors, effectively balances perceptual quality and distortion, enhancing generated image qualities.
% This training-sampling mismatch is akin to the ``exposure bias" observed in autoregressive models for text generation, where the models are trained on actual ground-truth sequences but transition to operating on their previous predicted words during inference \cite{bengio2015scheduled, RanzatoCAZ16Sequence, rennie2017self, Schmidt19closer}. Moreover, 
% \newpage

\section{Method}
\label{sec:method}

\begin{figure*}
    \centering
\includegraphics[width=.98\textwidth]{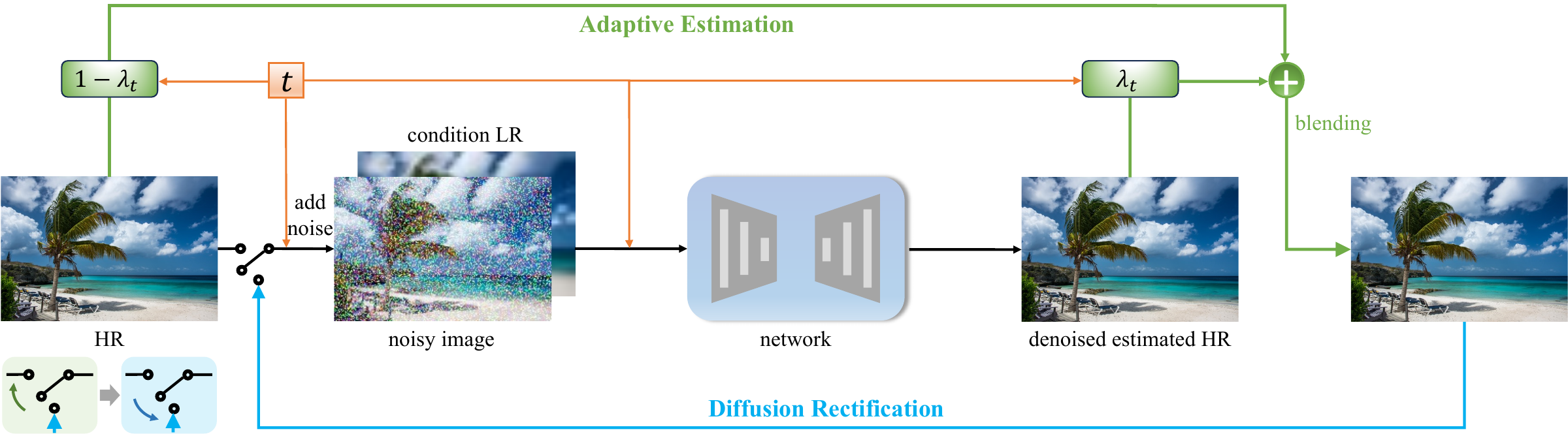}
    \caption{\textbf{Overview of the DREAM framework}. Starting with ground-truth HR images,  a standard diffusion process with a frozen denoiser network generates denoised HR estimates. The \textcolor{pipgreen}{Adaptive Estimation} merges these estimated HR images with the original HR images, guided by the pattern of estimation errors.  The \textcolor{pipblue}{Diffusion Rectification} constructs the noisy images from this merged HR images,  which are then fed into the denoiser network (now unfrozen). Similar to DDPM~\cite{ho2020denoising}, the denoiser network is trained to eliminate both the introduced Gaussian noise and errors arising from the training-sampling discrepancy, as detailed in~\cref{eq:dream-objective}.
   % \zz{The figure is very clear, except for the input to the noisy image. Maybe add a switch to distinguish the two states?} 
   }
    \label{fig:overview}
    \vspace{-.1in}
\end{figure*}

% \td{We first provide the preliminaries (\Cref{subsec:3.1}), following which we discuss the challenges associated with the training-sampling discrepancy observed in existing diffusion processes (\Cref{subsec:3.2}). Then, we present DreamSR, a novel diffusion model augmented with dynamic rectification (\Cref{subsec:3.3}) and adaptive estimation (\Cref{subsec:3.4}) tailored for the super-resolution (SR) task, as illustrated in~\Cref{fig:overview}. The training objective is given in~\Cref{subsec:3.5}.}

\subsection{Preliminaries}\label{subsec:3.1}
% \subsection{Background}

The goal of SR is to recover a high-resolution (HR) image from its low-resolution (LR) counterpart. This task is recognized as ill-posed due to its one-to-many nature~\cite{saharia2022image,yue2023resshift}, and is further complicated by various degradation models in real-world scenarios. Notably, diffusion models~\cite{sohl2015deep, ho2020denoising} have emerged as powerful generative models, showcasing strong capabilities in image generation tasks. Following~\cite{saharia2022image}, we address the SR challenge by adapting a \emph{conditional} denoising diffusion probabilistic
(DDPM) model. This adaptation, conditioned on the LR image, sets it apart from traditional, unconditional models which are primarily designed for unconstrained image generation.

We denote the LR and HR image pair as $(\vx_0, \vy_0)$. A conditional DDPM model involves a Markov chain, encompassing a \emph{forward process} that traverses the chain, adding noise to $\vy_0$, and a \emph{reverse process}, which conducts reverse sampling from the chain for denosing from pure Gaussain noise to the HR image $\vy_0$, conditioned on the LR image $\vx_0$.

\begin{algorithm}[t]
\small
\caption{Conditional DDPM Training}
\begin{algorithmic}[1]
    \REPEAT
    \STATE $(\vx_0, \vy_0)\sim p(\vx_0, \vy_0), t\sim\sU(1,T), \bm\rvepsilon_t\sim\gN(\vzero, \mI)$
    \STATE Compute $\vy_t = \sqrt{\bar{\alpha}_t}\vy_0 + \sqrt{1-\bar{\alpha}_t}\bm\rvepsilon_t$ \label{line:update-yt-training}
    \STATE
    % Take gradient descent step on
    Update $\theta$ with gradient $\nabla_\rvtheta||\bm\rvepsilon_t-\bm\rvepsilon_\rvtheta(\vx_0,\vy_t, t)||_1$
    \UNTIL{converged}
\end{algorithmic}
% \vspace{-.1in}
\label{sr3-training}
\end{algorithm}

\textbf{Forward process.} The forward process, also referred to as the diffusion process, takes a sample $\vy_0$ and simulates the non-equilibrium thermodynamic diffusion process~\cite{sohl2015deep}. It gradually adds Gaussian noise to $\vy_0$ via a fixed Markov chain of length $T$:

\vspace{-.25in}
\begin{align}
\small
q(\vy_t|\vy_{t-1}) &= \gN(\vy_t; \sqrt{1-\beta_t}\vy_{t-1}, \beta_t \mI),\\
q(\vy_{1:T}| \vy_0) &=\prod_{t=1}^Tq(\vy_t|\vy_{t-1}),
\end{align}
\vspace{-.15in}

\noindent
where $\{\beta_t\in (0,1)\}_{t=1}^T$ is the variance scheduler. As the step $t$ increases, the signal $\vy_0$ gradually loses its distinguishable features. Ultimately, as $t\to\infty$, $\vy_t$ converges to an isotropic Gaussian distribution. Moreover, we can derive the distribution for sampling at arbitrary step $t$ from $\vy_0$:

\vspace{-.2in}
\begin{align}
\small
    q\left(\vy_t | \vy_0\right)=\mathcal{N}\left(\vy_t ; \sqrt{\bar{\alpha}_t} \vy_0,\left(1-\bar{\alpha}_t\right) \boldsymbol{I}\right) .
    \label{eq:forward-yt}
\end{align}
where $\bar{\alpha}_t=\prod_{i=1}^t\alpha_i$ and $\alpha_t=1-\beta_t$.

\begin{algorithm}[t]
\small
\caption{Conditional DDPM Sampling}
\begin{algorithmic}[1]
    \STATE ${\vy}_T\sim\gN(\vzero, \mI)$
\FOR{$t=T \cdots 1$}
\STATE $\vz\sim\gN(\vzero,\mI)$ if $t>1$ else $\vz=\vzero$
\STATE ${\vy}_{t-1}=\frac{1}{\sqrt{\alpha_t}}({\vy}_t-\frac{1-\alpha_t}{\sqrt{1-\bar{\alpha}_t}}\bm\rvepsilon_\rvtheta(\vx_0, {\vy}_t, t))+\sigma_t\vz$\label{line:update-yt-sampling}
\ENDFOR
\STATE return ${\vy}_0$
\end{algorithmic}
\label{sr3-sampling}
\end{algorithm}

\textbf{Reverse process.} The reverse process, also referred to as the denosing process,  learns the conditional distributions $ p_\theta(\vy_{t-1} | \vy_t, \vx_0) $ for denoising from Gaussian noise to  $\vy_0$ conditioned on $\vx_0$, through a reverse Markovian process:

\vspace{-.2in}
\begin{align}
\small
    p_\rvtheta(\vy_{t-1}|\vy_{t}, \vx_0) &= \gN(\vy_{t-1}; \vmu_\rvtheta(\vx_0,\vy_t,t), \sigma_t^2\mI),\label{eq:reverse-yt}\\
p_\rvtheta(\vy_{0:T}| \vx_0)&=p(\vy_T)\prod_{t=1}^Tp_\rvtheta(\vy_{t-1}|\vy_{t},\vx_0),
\end{align}
\vspace{-.2in}

\noindent
where $\sigma_t$ is a predetermined term related to $\beta_t$~\cite{ho2020denoising}.

\textbf{Training.} We train a denoising network $\bm\epsilon_\theta(\bm x_0,\vy_t, t)$ to predict the noise vector $\bm\epsilon_t$ added at step $t$. Following~\cite{ho2020denoising, saharia2022image}, the training objective can be expressed as:

\vspace{-.2in}
\begin{align}
\small
    \gL(\rvtheta) = \E_{(\vx_0,\vy_0), \bm\rvepsilon_t, t}\left\|\bm\rvepsilon_t - \bm\rvepsilon_\rvtheta(\vx_0,\vy_t, t)\right\|_1.
    \label{eq:dm-objective}
\end{align}
\vspace{-.2in}

\noindent
With~\cref{eq:forward-yt}, we parameterize $\vy_t=\sqrt{\bar{\alpha}_t}\vy_0 + \sqrt{1-\bar{\alpha}_t}\bm\rvepsilon_t$, and summarize the training process in~\cref{sr3-training}.

\textbf{Sampling.} In essence, the training minimizes the divergence between the forward posterior $q\left(\vy_{t-1} | \vy_t, \vy_0\right)$ and $p_\theta\left(\vy_{t-1} | \vy_t,\vx_0\right)$, and the mean $\vmu_\rvtheta(\vx_0, \vy_t, t)$ in~\cref{eq:reverse-yt} is parameterized~\cite{saharia2022image} to match the mean of $q\left(\vy_{t-1} | \vy_t, \vy_0\right)$:

\vspace{-.2in}
\begin{align}\label{eq:sampling-mean}
\small
    \vmu_\rvtheta(\vx_0,\vy_t, t) = \frac{1}{\sqrt{\alpha_t}}(\vy_t-\frac{1-\alpha_t}{\sqrt{1-\bar{\alpha}_t}}\bm\rvepsilon_\rvtheta(\vx_0,\vy_t, t)).
\end{align}
\vspace{-.15in}

\noindent
To sample $\vy_0\sim p_\theta(\vy_0| \vx_0)$, starting from $\vy_T\sim\mathcal{N}(\bm0, \bm I)$, we reverse the Markovian process by iteratively sampling $\vy_{t-1}\sim p(\vy_{t-1}| \vy_{t}, \vx_0)$ based on~\cref{eq:reverse-yt,,eq:sampling-mean}, which completes the sampling process, as shown in~\cref{sr3-sampling}.

\subsection{Challenge: training-sampling discrepancy}
\label{subsec:3.2}

Training diffusion models for SR presents a critical challenge, stemming from a discrepancy between the training and inference phases, which we term as \emph{training-sampling discrepancy}. During the training phase, the model operates on actual data, wherein the noisy image $\vy_t$ at diffusion step $t$ is derived from the \emph{ground-truth} HR image $\vy_0$ as per line~\ref{line:update-yt-training} in~\cref{sr3-training}. However, during the inference phase, the ground truth $\vy_0$ is unavailable. As outlined in line~\ref{line:update-yt-sampling} in~\cref{sr3-sampling}, the model now operates on predicted data, where $\vy_t$ is obtained from the preceding sampling step $t+1$. Due to the estimation error, the noisy image ${\vy}_t$ constructed in these two processes usually differs, giving rise to the training-sampling discrepancy.

To better illustrate the discrepancy, we conduct an experiment utilizing a pre-trained SR3 model~\cite{saharia2022image}, denoted by $\bm\epsilon_\theta$, adhering to the standard diffusion training framework. The goal is to understand the implications for HR signal $\vy_0$ reconstruction under two distinct scenarios:
\begin{itemize}%[leftmargin=.1in]
    \item ``Training". Simulating the training process, we \emph{assume access} to the ground-truth $\vy_0$, and construct the noisy image at time step $t$ as per line~\ref{line:update-yt-training} in~\cref{sr3-training}, denoting the image as $\vy_t^{\text{train}}$.
    \item ``Sampling". Simulating the sampling process, we \emph{assume no access} to $\vy_0$ and iteratively construct the noisy image at each time step $t$ by sampling from the previous step, as per line~\ref{line:update-yt-sampling} in~\cref{sr3-sampling}. The noisy image thus obtained is denoted by $\vy_t^{\text{sample}}$.
\end{itemize}

To retrieve the HR image $\vy_0$ from the noisy image in both scenarios, we utilize~\cref{eq:forward-yt} and the pre-trained network $\bm\epsilon_\theta$ to compute the predicted HR signal as follows:

\vspace{-.2in}
\begin{equation}
\small
\widetilde{\boldsymbol{y}}_0=\frac{1}{\sqrt{\bar{\alpha}_t}}\left(\boldsymbol{y}_t-\sqrt{1-\bar{\alpha}_t} \bm\epsilon_\theta\left(\boldsymbol{x}_0, \boldsymbol{y}_t, t\right)\right)=:h_\theta(\vy_t).
\label{eq:estimate_y0}
\end{equation}
\vspace{-.15in}

\noindent
Following this, we compute $\widetilde{\vy}_0^{\text{train}}=h_\theta(\vy_t^{\text{train}})$ and $\widetilde{\vy}_0^{\text{sample}}=h_\theta(\vy_t^{\text{sample}})$ as the predicted HR images in the ``training" and ``sampling" scenarios, respectively. For performance evaluation, we take 100 samples from FFHQ~\cite{karras2019style} and calculate the averaged MSE and LPIPS~\cite{zhang2018unreasonable} metrics between the predicted HR images and the ground-truth $\vy_0$ across various time step $t$ under the defined settings.

\begin{figure} [t]
     \centering
     \begin{subfigure}[b]{0.23\textwidth}
         \centering
         \includegraphics[width=\textwidth]{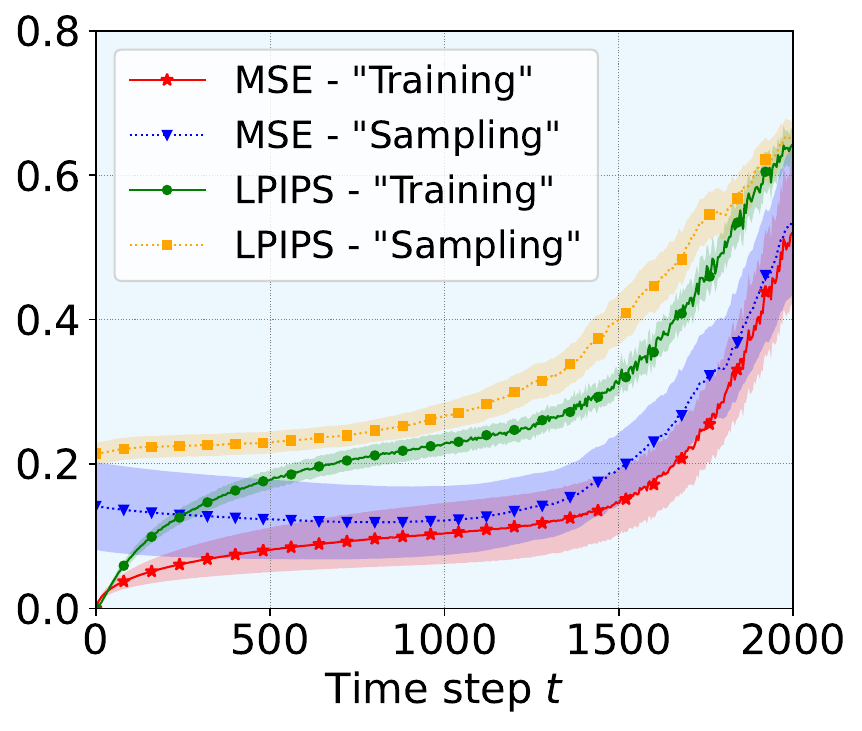}
         \caption{Standard diffusion}
         \label{fig:pixel-error-dynamic}
     \end{subfigure}
     % \qquad\qquad
    % \hfill
     \begin{subfigure}[b]{0.23\textwidth}
         \centering
         \includegraphics[width=\textwidth]{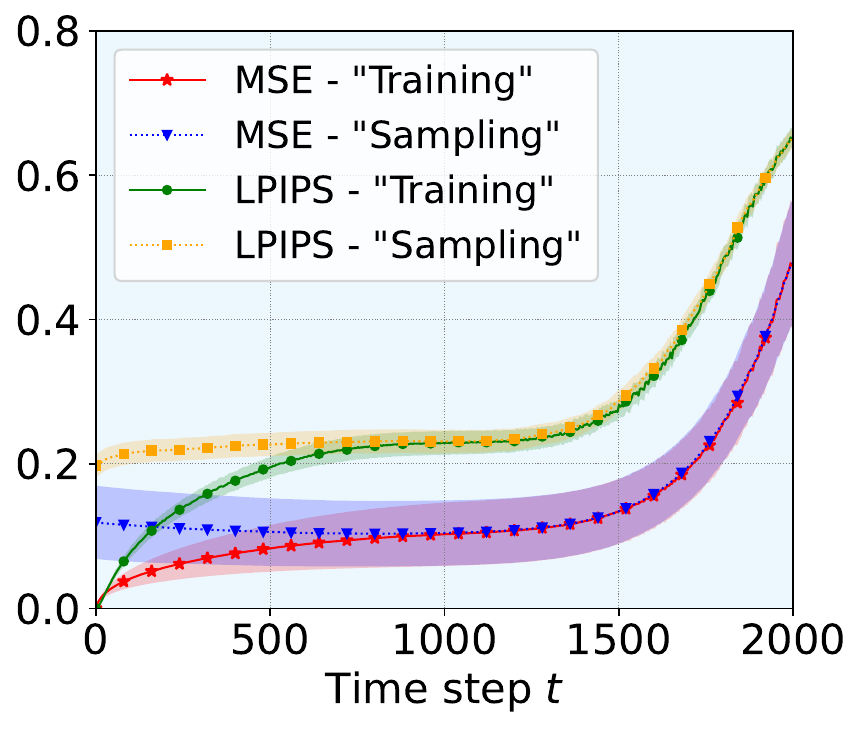}
         \caption{DREAM}
         \label{fig:latent-error-dynamic}
     \end{subfigure}\
     \vspace{-.1in}
         \caption{Evaluation of training-sampling discrepancy and its alleviation through our DREAM framework. The mean curve over 100 samples at each time step $t$ is plotted, with the shaded area representing the standard deviation of each metric.  Here, $T=2000$.}
        \label{fig:error-dynamic}
        \vspace{-.1in}
\end{figure}

We present the findings in~\Cref{fig:pixel-error-dynamic}, where both MSE and LPIPS exhibit a decline with a smaller $t$, as expected, since the network can reconstruct more accurate HR signal from less noisy input. Importantly, discernible disparities are observed between the curves representing the ``training" and ``sampling" settings---the ``training" curves consistently exhibit lower error compared to the ``sampling" ones, suggesting the advantage of having access to the ground-truth $\vy_0$ for improved prediction accuracy. In contrast, \Cref{fig:latent-error-dynamic} illuminates a remarkable alleviation in this discrepancy when employing our DREAM framework to train the identical SR3 architecture: \emph{the ``sampling" curve closely aligns with the ``training" curve, despite the lack of access to the ground-truth $\vy_0$, across both MSE and LPIPS metrics.} This underscores the efficacy of our approach in bridging the training-sampling discrepancy and thereby facilitating more accurate predictions.
\subsection{The DREAM framework} \label{subsec:3.3}

We now present our DREAM framework (see~\Cref{fig:overview}), an end-to-end training strategy designed to bridge the gap between training and sampling in diffusion models. It consists of two core components: \emph{diffusion rectification} and \emph{estimation adaptation}, which we elaborate as follows.
% In this part, we introduce a two-phase end-to-end training strategy, composed of two crucial components: \emph{dyna rectification} and \emph{estimation adaptation}, which can effectively align the training process with the sampling process.  
% \label{subsec:3.3}
% \jz{not sure where to introduce the SA in the depth estimation.}

\begin{figure}
    \centering
    \includegraphics[width=0.99\columnwidth]{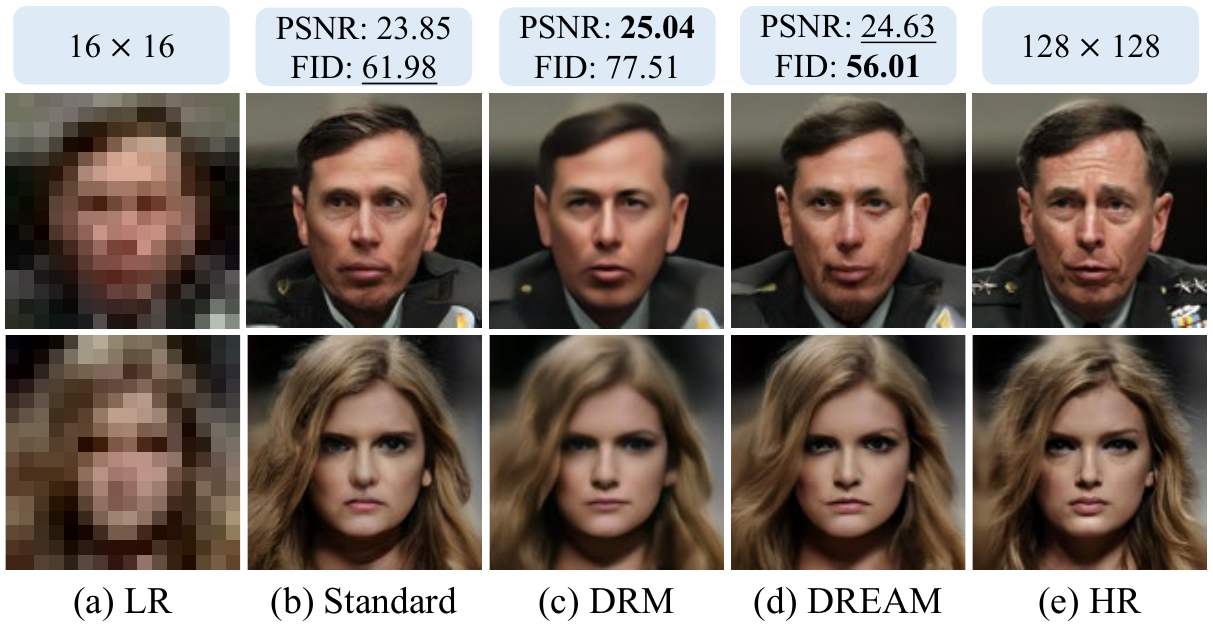}
    \vspace{-.1in}
    \caption{
    % Qualitative comparison on 8$\times$ SR on CelebA-HQ \cite{karras2017progressive} between standard and DRM methods. 
    $8\times$ SR on the CelebA-HQ dataset~\cite{karras2017progressive}.} 
        \label{fig:drm-visual}
    \vspace{-.1in}
\end{figure}

\textbf{Diffusion rectification.} The goal of diffusion rectification is to modify the behavior of the diffusion training to account for the training-sampling discrepancy, which arises from the manner in which we construct the intermediate signals---either from the ground-truth or from the model's own estimation.  Hence, we extend the diffusion training framework to align more closely with the sampling process, enabling the model to utilize its own output for prediction.

Specifically, during training, upon acquiring $\vy_t^{\text{train}}$ as per line~\ref{line:update-yt-training} in~\cref{sr3-training}, we refrain from directly  minimizing $\mathcal{L}(\theta)$. Instead,  we construct our own prediction of the HR image as $\widetilde{\vy}_0^{\text{train}}$ according to~\cref{eq:estimate_y0}, formulated as:

\vspace{-.2in}
\begin{align}\label{eq:y0-train-drm}
\small
    \begin{split}
    \widetilde{\boldsymbol{y}}_0^{\text{train}}&=\frac{1}{\sqrt{\bar{\alpha}_t}}\big(\boldsymbol{y}_t^{\text{train}}-\sqrt{1-\bar{\alpha}_t} \bm\epsilon_\theta(\boldsymbol{x}_0, \boldsymbol{y}_t^{\text{train}}, t)\big)\\
        &=\frac{1}{\sqrt{\bar{\alpha}_t}}\big( \sqrt{\bar{\alpha}_t}\vy_0 + \sqrt{1-\bar{\alpha}_t}\bm\rvepsilon_t    \hspace{.48in}\triangleright\text{line~\ref{line:update-yt-training}}   \\
        &\hspace{1.1in}-\sqrt{1-\bar{\alpha}_t} \bm\epsilon_\theta(\boldsymbol{x}_0, \boldsymbol{y}_t^{\text{train}}, t)\big)\\
        &=\vy_0 + \sqrt{(1-\bar{\alpha}_t)/{\bar{\alpha}_t}} \Delta\bm{\rvepsilon}_{t,\theta}
    \end{split}
\end{align}
\vspace{-.15in}

\noindent
where $\Delta\bm{\rvepsilon}_{t,\theta} = \bm{\rvepsilon}_t-\bm{\rvepsilon}_\rvtheta(\vx_0,\vy_t^{\text{train}},t)$. Utilizing this self-estimated HR image $\widetilde{\vy}_0^{\text{train}}$,  we generate the noisy image $\widetilde{\vy}_t^{\text{train}}$ to serve as input\footnote{To match the actual sampling process, there might be a desire to reconstruct $ \widetilde{\vy}_{t-1}^{\text{train}}$, yet this could notably complicate the entire procedure. Nonetheless, we have observed similar performance by simply using $ \widetilde{\vy}_t^{\text{train}}$.} to the network $\bm\epsilon_\theta$ once more:

\vspace{-.2in}
\begin{align}\label{eq:self-supervise}
\small
\begin{split}
    \widetilde{\vy}_t^{\text{train}} &= \sqrt{\bar{\alpha}_t}\widetilde{\vy}_0^{\text{train}}+\sqrt{1-\bar{\alpha}_t}\bm\epsilon'_t\\
    &= \sqrt{\bar{\alpha}_t}\vy_0+ \sqrt{1-\bar{\alpha}_t}(\bm \epsilon'_t+\Delta\bm{\rvepsilon}_{t,\theta}),
\end{split}
\end{align}
\vspace{-.15in}

\noindent
where $\bm\epsilon'_t\sim\mathcal{N}(\bm0, \bm I)$. Then, the training objective for this diffusion rectification model (DRM) can be expressed as:

\vspace{-.25in}
\begin{align}\label{eq:drm-objective}
\small
    \gL^{\text{DRM}}(\rvtheta) = \E_{(\vx_0,\vy_0), \bm\rvepsilon_t,\bm\rvepsilon'_t, t} \left\|\big(\bm\rvepsilon'_t +\Delta\bm{\rvepsilon}_{t,\theta}\big) -\bm\rvepsilon_\rvtheta(\vx_0,\widetilde{\vy}_t^{\text{train}}, t) \right\|_1.
\end{align}
\vspace{-.2in}

Essentially,~\cref{eq:drm-objective} suggests that this DRM approach strives not only to eliminate the sampled noise $\bm\rvepsilon'_t$  but also to address the error term $\Delta\bm{\rvepsilon}_{t,\theta}$ arising from the discrepancy between the imperfect estimation $\widetilde{\vy}_0^{\text{train}}$ and the ground-truth $\vy_0$, as seen in~\cref{eq:y0-train-drm}; hence the term ``rectification". Notably, leveraging the model's own prediction during training as in~\cref{eq:self-supervise} mirrors the sampling process of DDIM~\cite{song2021denoising} with a particular choice of $\sigma_t$, thereby imposing enhanced supervision. We remark that DRM is closely related to the approaches in~\cite{saxena2023monocular,savinov2022stepunrolled,ji2023ddp} where they perform similar step-unrolling techniques for perceptual vision tasks or text generation tasks. However, we are the first to tailor it to low-level vision tasks and provide a clear analysis.

\begin{algorithm}[t]
\small
\caption{Conditional DREAM Training}
\begin{algorithmic}[1]\label{alg:dr-training}
    \REPEAT
    \STATE $(\vx_0, \vy_0)\sim p(\vx_0, \vy_0), t\sim\sU(1,T), \bm\rvepsilon_t\sim\gN(\vzero, \mI)$
    \STATE Compute $\vy_t = \sqrt{\bar{\alpha}_t}\vy_0 + \sqrt{1-\bar{\alpha}_t}\bm\rvepsilon_t$
    \STATE Compute $\Delta\bm{\rvepsilon}_{t,\theta}=\bm{\rvepsilon}_t-\texttt{StopGradient}(\bm{\rvepsilon}_\rvtheta(\vx_0, \vy_t, t))$ \label{line:start}
    \STATE
    Compute $\widehat{\vy}_t=\vy_t+ \sqrt{1-\bar{\alpha}_t}\lambda_t\Delta\bm{\rvepsilon}_{t,\theta}$
    \STATE Update $\theta$ with gradient  $\nabla_\rvtheta||\bm\rvepsilon_t +\lambda_t\Delta\bm{\rvepsilon}_{t,\theta}-\bm\rvepsilon_\rvtheta(\vx_0,\widehat{\vy}_t, t)||_1$\label{line:end}
    \UNTIL{converged}
\end{algorithmic}
\end{algorithm}

\textbf{Estimation adaptation.} While DRM incorporates additional rectification supervision to account for the sampling process, its naive application to the SR task might not deliver satisfactory results. As shown in~\Cref{fig:drm-visual}, a distortion-perception tradeoff~\cite{blau2018perception} is observed in the generated SR images.  Despite achieving a state-of-the-art PSNR (less distortion), the images produced by DRM tend to be smoother and lack fine details, reflecting a high FID score (poor perception). This is particularly evident when compared to the standard conditional diffusion model, namely SR3~\cite{saharia2022image}.  This limitation could be traced back to DRM's static self-alignment mechanism, which may inappropriately guide the generated images to regress towards the mean.

To address the issue, and inspired by the powerful generative capability of the standard diffusion model, we propose an estimation adaptation strategy. This aims to harness both the superior quality of standard diffusion and the reduced distortion offered by diffusion rectification. Specifically, rather than naively using our own prediction $\widetilde{\vy}_0^\text{train}$ computed in~\cref{eq:y0-train-drm}, we adaptively inject ground-truth information $\vy_0$ by blending it with $\widetilde{\vy}_0^\text{train}$ as follows: 

\vspace{-.23in}
\begin{align}\label{eq:blend-hat-y0}
\small
    \widehat{\vy}_0 = \lambda_t \widetilde{\vy}_0^{\text{train}} + (1-\lambda_t) \vy_0,
\end{align}
\vspace{-.22in}

\noindent
where $\lambda_t\in(0,1)$ is an increasing function such that $\widehat{\vy}_0$ emphasizes more on $\vy_0$ at smaller $t$, aligning with the network's tendency to achieve more accurate predictions, as observed in~\Cref{fig:error-dynamic}. Intuitively, as $t$ decreases, $\widehat{\vy}_0$ closely approximates the ground-truth, making it more beneficial to resemble the standard diffusion, yielding images with realistic details. Conversely, as $t$ increases and the prediction leans towards random noise, it is advantageous to focus more on the estimation itself, effectively aligning the training and sampling processes through the rectification.

\begin{table*}[t]
\centering
\caption{Comparison  on face and general scene datasets against three baselines for various $p$ values, with  \colorbox{red!20}{best} and \colorbox{orange!20}{second-best} colorized.}
\vspace{-.1in}
\label{tab:face-sr3-peffect}
\footnotesize
\tabcolsep=0.30em
\begin{tabular}{ccccccccc|cccccccc}
\toprule
\multirow{3}{*}{$p$} & \multicolumn{8}{c|}{CelebA-HQ~\cite{karras2017progressive}} & \multicolumn{8}{c}{DIV2K~\cite{agustsson2017ntire}} \\ \cmidrule(lr){2-17} 
 & \multicolumn{4}{c|}{SR3~\cite{saharia2022image}} & \multicolumn{4}{c|}{IDM~\cite{gao2023implicit}} & \multicolumn{4}{c|}{SR3~\cite{saharia2022image}} & \multicolumn{4}{c}{ResShift~\cite{yue2023resshift}} \\ %\cline{2-17} 
 & PSNR$\uparrow$ & SSIM$\uparrow$ & LPIPS$\downarrow$ & \multicolumn{1}{c|}{FID$\downarrow$} & PSNR$\uparrow$ & SSIM$\uparrow$ & LPIPS$\downarrow$ & \multicolumn{1}{c|}{FID$\downarrow$} & PSNR$\uparrow$ & SSIM$\uparrow$ & LPIPS$\downarrow$ & \multicolumn{1}{c|}{FID$\downarrow$} & PSNR$\uparrow$ & SSIM$\uparrow$ & LPIPS$\downarrow$ & FID$\downarrow$ \\ \midrule
\multicolumn{1}{l}{$0$ (DRM)} & \cellcolor{red!20}{$25.04$} & \cellcolor{red!20}$0.76$ & $0.204$ & \multicolumn{1}{c|}{$77.51$} & \cellcolor{red!20}$25.06$ & \cellcolor{red!20}$0.76$ & $0.188$ & $67.46$ & \cellcolor{red!20}$28.67$ & \cellcolor{red!20}$0.81$ & $0.189$ & \multicolumn{1}{c|}{$16.72$} & \cellcolor{red!20}$29.98$ & \cellcolor{red!20}$0.83$ & $0.233$ & $17.76$ \\\midrule
\multicolumn{1}{l}{$1$ (DREAM)} & \cellcolor{orange!20}$24.63$ & \cellcolor{orange!20}$0.74$ & \cellcolor{red!20}$0.177$ & \multicolumn{1}{c|}{ \cellcolor{red!20}$56.01$} & \cellcolor{orange!20}$24.50$ & \cellcolor{orange!20}$0.73$ & \cellcolor{red!20}$0.167$ & \cellcolor{red!20}$53.22$ & \cellcolor{orange!20}$28.10$ & \cellcolor{orange!20}$0.79$ & \cellcolor{red!20}$0.121$ & \multicolumn{1}{c|}{\cellcolor{red!20}$14.32$} & \cellcolor{orange!20}$29.24$ & \cellcolor{orange!20}$0.80$ & $0.158$ & $16.23$ \\
\multicolumn{1}{l}{$2$ (DREAM)} & $24.62$ & \cellcolor{orange!20}$0.74$ & \cellcolor{orange!20}$0.180$ & \multicolumn{1}{c|}{$61.72$} & $24.32$ & $0.72$ & \cellcolor{orange!20}$0.169$ & $55.38$ & $28.06$ & \cellcolor{orange!20}$0.79$ & $0.140$ & \multicolumn{1}{c|}{$15.54$} & $28.77$ & $0.79$ & \cellcolor{orange!20}$0.134$ & \cellcolor{orange!20}$15.72$ \\
\multicolumn{1}{l}{$3$ (DREAM)} & $24.15$ & $0.71$ & $0.182$ & \multicolumn{1}{c|}{ \cellcolor{orange!20}$58.89$} & $24.09$ & $0.72$ & $0.172$ & \cellcolor{orange!20}$54.04$ & $27.88$ & \cellcolor{orange!20}$0.79$ & \cellcolor{orange!20}$0.123$ & \multicolumn{1}{c|}{\cellcolor{orange!20}$14.83$} & $28.44$ & $0.79$ & \cellcolor{red!20}$0.124$ & \cellcolor{red!20}$15.67$ \\\midrule
\multicolumn{1}{l}{$\infty$ (standard)} & $23.85$ & $0.71$ & $0.184$ & \multicolumn{1}{c|}{$61.98$} & $24.01$ & $0.71$ & $0.172$ & $56.01$ & $27.02$ & $0.76$ & \cellcolor{red!20}$0.121$ & \multicolumn{1}{c|}{$16.72$} & $25.30$ & $0.68$ & $0.211$ & $25.91$ \\ \bottomrule
\end{tabular}
\vspace{-.1in}
\end{table*}

Following the adaptive estimation $\widehat{\vy}_0$ in~\cref{eq:blend-hat-y0}, we construct the new noisy image $\widehat{\vy}_t$ similarly as before:

\vspace{-.2in}
\begin{align}\label{eq:hat-yt}
\begin{split}
\small
    \widehat{\vy}_t & = \sqrt{\bar{\alpha}_t}\widehat{\vy}_0+\sqrt{1-\bar{\alpha}_t}\bm \epsilon'_t \\
    &= \sqrt{\bar{\alpha}_t}\vy_0+ \sqrt{1-\bar{\alpha}_t}(\bm\epsilon'_t+\lambda_t\Delta\bm{\rvepsilon}_{t,\theta}).
\end{split}
\end{align}
\vspace{-.1in}

\noindent
Finally, the training objective for our full \emph{Diffusion Rectification and Estimation-Adaptive Model (DREAM)} can be expressed as:

\vspace{-.2in}
\begin{align}\label{eq:dream-objective}
\small
    \gL^{\text{DREAM}}(\rvtheta) = \E_{(\vx_0,\vy_0), \bm\rvepsilon_t,\bm\rvepsilon'_t, t} \left\|\big(\bm\rvepsilon'_t +\lambda_t\Delta\bm{\rvepsilon}_{t,\theta}\big) -\bm\rvepsilon_\rvtheta(\vx_0,\widehat{\vy}_t, t) \right\|_1.
\end{align}
\vspace{-.2in}

\textbf{Choice of $\lambda_t$.} Comparing~\cref{eq:dream-objective} with~\cref{eq:drm-objective}, the key difference lies in the introduction of $\lambda_t$ for adaptively modulating the intensity of the rectification term $\Delta\bm\epsilon_{t,\theta}$. Note that we only need $\lambda_t\in(0,1)$ to be increasing to leverage the benefits of both standard diffusion and rectification. In practice, we set $\lambda_t=(\sqrt{1-\bar{\alpha}_t})^{p}$, where $p$ adds an extra layer of flexibility: at $p=0$, $\lambda_t$ remains at 1, reverting the method to DRM with consistent static rectification; as $p\to\infty$, $\lambda_t\to0$, transitioning our approach towards the standard diffusion model. As shown in~\Cref{fig:drm-visual}, the images produced by DERAM with $p=1$ achieve a superior balance between perception and distortion, significantly outperforming the standard SR3~\cite{saharia2022image} across both metrics.

\textbf{Training details.}  It's important to highlight that while the same network $\bm\epsilon_\theta$ is utilized for calculating both the rectification term $\Delta\bm\epsilon_{t,\theta}$ and the predicted noise $\bm\rvepsilon_\rvtheta(\vx_0,\widehat{\vy}_t, t)$ in~\cref{eq:dream-objective}, a key distinction exists:  we refrain from propagating the gradient when computing $\Delta\bm\epsilon_{t,\theta}$, and thus, it is derived from the frozen network. The actual supervision is imposed following its adaptive adjustment. Moreover, we empirically observe that using the same Gaussian noise (\ie, $\bm{\rvepsilon}_t\equiv\bm{\rvepsilon}'_t$) in DREAM yields superior performance, further simplifying~\cref{eq:hat-yt} to:
\begin{align}
    \widehat{\vy}_t & = \vy_t^{\text{train}}+ \sqrt{1-\bar{\alpha}_t}\lambda_t\Delta\bm{\rvepsilon}_{t,\theta}.
\end{align}
We summarize our DREAM framework in~\Cref{alg:dr-training}, tailored for enhanced diffusion training, while~\cref{sr3-sampling} remains applicable for sampling purposes.
\section{Experiments}
\label{sec:exp}

% Beyond the comprehensive experiments presented in this section, additional results and analysis are available in the supplementary materials.

\subsection{Implementation details}

\textbf{Baselines and datasets.}
Our experiments involve three diffusion-based SR methods as baselines, spanning datasets for faces, general scenes, and natural images. For face image datasets, we adopt SR3\footnote{Due to the unavailability of official code, we use a widely-recognized implementation \href{https://github.com/Janspiry/Image-Super-Resolution-via-Iterative-Refinement}{[link]}.}~\cite{saharia2022image}  and IDM~\cite{gao2023implicit} as baselines, with  training conducted on FFHQ~\cite{karras2019style} and evaluations on CelebA-HQ~\cite{karras2017progressive}.  For general scenes, we use the DIV2K dataset~\cite{agustsson2017ntire}, employing SR3~\cite{saharia2022image} and ResShift\footnote{To ensure consistency across baselines, we standardize the transition kernel to align with DDPM's approach for noise prediction.}~\cite{yue2023resshift} as baseline models. Notably, SR3 and IDM operate in pixel space, whereas ResShift conducts diffusion process in latent space. In addition, to assess out-of-distribution (OOD) performance, we train SR3 as baseline on the DIV2K dataset and evaluate on CAT~\cite{zhang2008cat} and LSUN datasets~\cite{yu2015lsun}. 
% More implementation details can be found in the appendix.
% Our experiments are carried out using three distinct baselines across both face datasets, natural image datasets, and a general scene dataset (DIV2K \cite{agustsson2017ntire}). Specifically for face datasets, we adopt SR3~\cite{saharia2022image} \footnote{Because the official code is unavailable, our experiments adopt the popular implementation \href{https://github.com/Janspiry/Image-Super-Resolution-via-Iterative-Refinement}{[link]}.} and IDM~\cite{gao2023implicit} as our baseline models for comparison, adhering to the same settings of training on FFHQ ~\cite{karras2019style} and evaluation on CelebA-HQ ~\cite{karras2017progressive}. For general scenes, we adopt SR3~\cite{saharia2022image} and ResShift~\cite{yue2023resshift} as our baseline models for comparison on DIV2K dataset~\cite{agustsson2017ntire}. Finally, to evaluate the out-of-distribution (OOD) performance, we train on DIV2K dataset and evaluate on the LSUN~\cite{yu2015lsun} dataset. More implementation details can be found in the appendix.

\begin{table}[]\small
\centering
\caption{Quantitative comparison for 16$\times$16  to 128$\times$128 face super-resolution on CelebA-HQ~\cite{karras2017progressive}. Consistency measures the MSE $(\times 10^{-5})$ between LR and downsampled SR images.}
\vspace{-.1in}
\label{tab:face}
\begin{tabular}{lccc}
\hline
Method     & { PSNR$\uparrow$} & { SSIM$\uparrow$} & {Consistency$\downarrow$} \\ \hline
PULSE~\cite{menon2020pulse}     & 16.88                       & 0.44                        & 161.1                              \\
FSRGAN~\cite{chen2018fsrnet}   & 23.85                       & 0.71                        & 33.8                               \\
Regression~\cite{saharia2022image} & 23.96                       & 0.69                        & 2.71                               \\
SR3~\cite{saharia2022image}        & 23.85                       & 0.71                        & 2.33\\
IDM~\cite{gao2023implicit}     & 24.01                       & 0.71                        & 2.14  
\\ \hline
SR3~\cite{saharia2022image}+DREAM        & \cellcolor{red!20}24.63                       & \cellcolor{red!20}0.74                       & \cellcolor{orange!20}2.12\\
IDM~\cite{gao2023implicit}+DREAM     & \cellcolor{orange!20}24.50                       & \cellcolor{orange!20}0.73                        & \cellcolor{red!20}1.26  
\\ \hline
\end{tabular}
\vspace{-.1in}
\end{table}

\subsection{Results and analysis}

\textbf{Effect of $p$ in $\lambda_t$.} In DREAM implementation, we set $\lambda_t = (\sqrt{1-\bar{\alpha}_t})^p$, where $p$ manages the balance between ground-truth and self-estimation data as in~\cref{eq:blend-hat-y0}.  We conduct experiments with three baselines (SR3, IDM and ResShift) for $8\times$ face SR on CelebA-HQ and $4\times$ general scene SR on DIV2K at various $p$ settings, as shown in~\Cref{tab:face-sr3-peffect}. Baselines use the standard diffusion process ($p\to\infty$). For $p=0$ ($\lambda_t\equiv1$), corresponding to the DRM model in~\cref{eq:drm-objective}, there is a notable reduction in distortion (higher PSNR and SSIM), but at the cost of perceptual quality (lower LPIPS and FID), confirming our findings in~\Cref{fig:drm-visual}. Increasing $p$ to $1$ (our full DREAM approach) leads to a slight decrease in distortion but significantly improves the balance between distortion and perception. Further increase in $p$ shows continual distortion degradation, while perceptual quality initially improves then declines. \emph{DREAM demonstrates clear advantages over baseline models across all metrics.} We found $p=1$ yields the \emph{best overall performance} compared to other $p$ values and baselines, making it our choice for subsequent experiments.

% The DREAM strategy in \cref{eq:dream-objective},  integrates a technique for dynamically combining diffusion rectification with standard diffusion. For simplicity in practice, we set $\lambda_t = (\sqrt{1-\bar{\alpha}_t})^p$, where the hyperparameter $p$ allows for adjustable balancing of information sourced from actual data versus that from self-estimation during the Markov process.  For $p=1$, $\lambda_t=1$ through the diffusion process, degrading the method to DRM. On the other hand, as $p\rightarrow \infty$, $\lambda_t\rightarrow0$, transforming the method to the standard diffusion. In \Cref{tab:face-sr3-peffect}, we  we present a comparative analysis against three baseline models for $8\times$ face super-resolution and $4\times$ general  scene (DIV2K) dataset  across various $p$ settings. The outcomes are evaluated based on distortion quality, using PSNR and SSIM, and on perceptual quality, using LPIPS and FID. As the parameter $p$ increases, there is a continuous degradation in the distortion of the SR images, while perceptual quality initially shows improvement but subsequently experiences a decline. When $p$ lies within the interval  $[1,3]$, our method attains superior results in terms of both distortion reduction and perceptual enhancement. Therefore, in this work, we have chosen to set $p=1$. 

\begin{figure}[t]
    \centering
    \includegraphics[width=0.99\columnwidth]{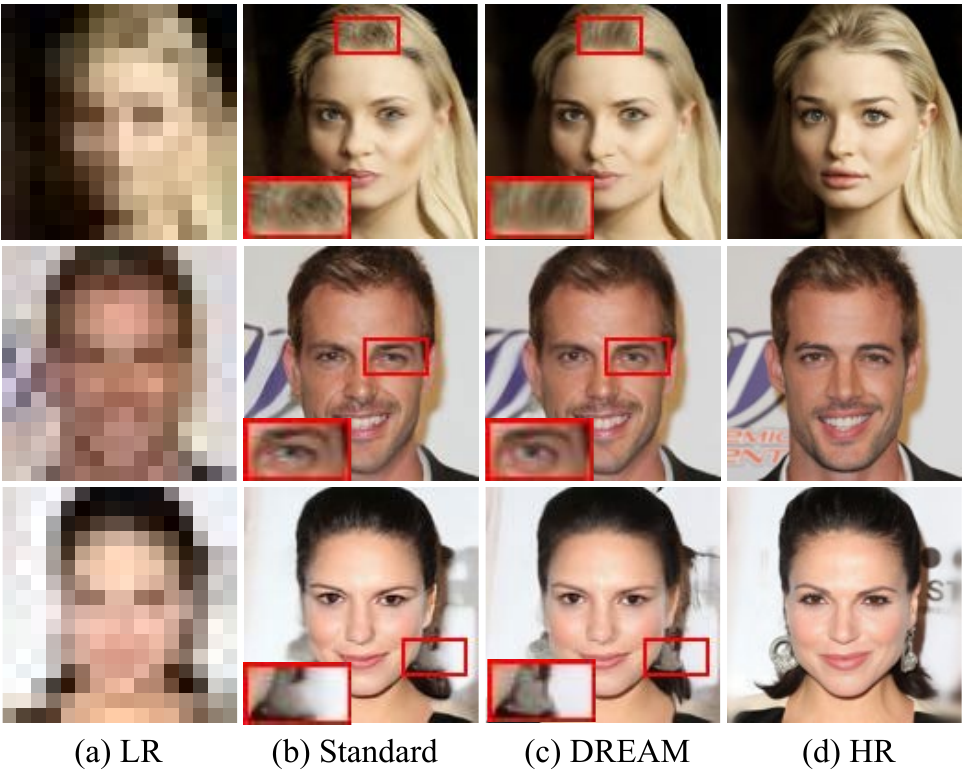}
    \vspace{-.1in}
    \caption{
    Qualitative comparison for $8\times$ SR using IDM~\cite{gao2023implicit} on the CelebA-HQ dataset~\cite{karras2017progressive}. Results highlight DREAM's superior fidelity and enhanced identity preservation, leading to more realistic detail generation in features like hair, eyes, and rings.} 
        \label{fig:face-idm}
    \vspace{-.1in}
\end{figure}

\textbf{Face super-resolution.}
\Cref{fig:drm-visual,,fig:face-idm} show qualitative comparisons for face super-resolution from $16\times16$ to $128\times128$, applying our DREAM approach to state-of-the-art diffusion-based methods, SR3 and IDM. While SR3 and IDM generally have decent image qualities, they often miss intricate facial details like hair and eyes, resulting in somewhat unrealistic appearance, and even omit accessories like rings. In contrast, our DREAM approach operated on the these baseline  more faithfully preserves facial identity and details. \Cref{tab:face} shows a quantitative comparison of our DREAM approach applied to SR3 and IDM against other methods, using metrics such as PSNR, SSIM, and consistency~\cite{saharia2022image}. While GAN-based models are known for their fidelity to human perception at higher SR scales, their lower consistency scores suggest a notable deviation from the original LR images. Applying DREAM to SR3 and IDM, we observe considerable enhancements across all metrics. Notably, the simpler SR3, a pure conditional DDPM, when augmented with DREAM, outperforms the more complex IDM, underscoring DREAM's effectiveness.

\begin{figure}[t]
    \centering
    \includegraphics[width=0.99\columnwidth]{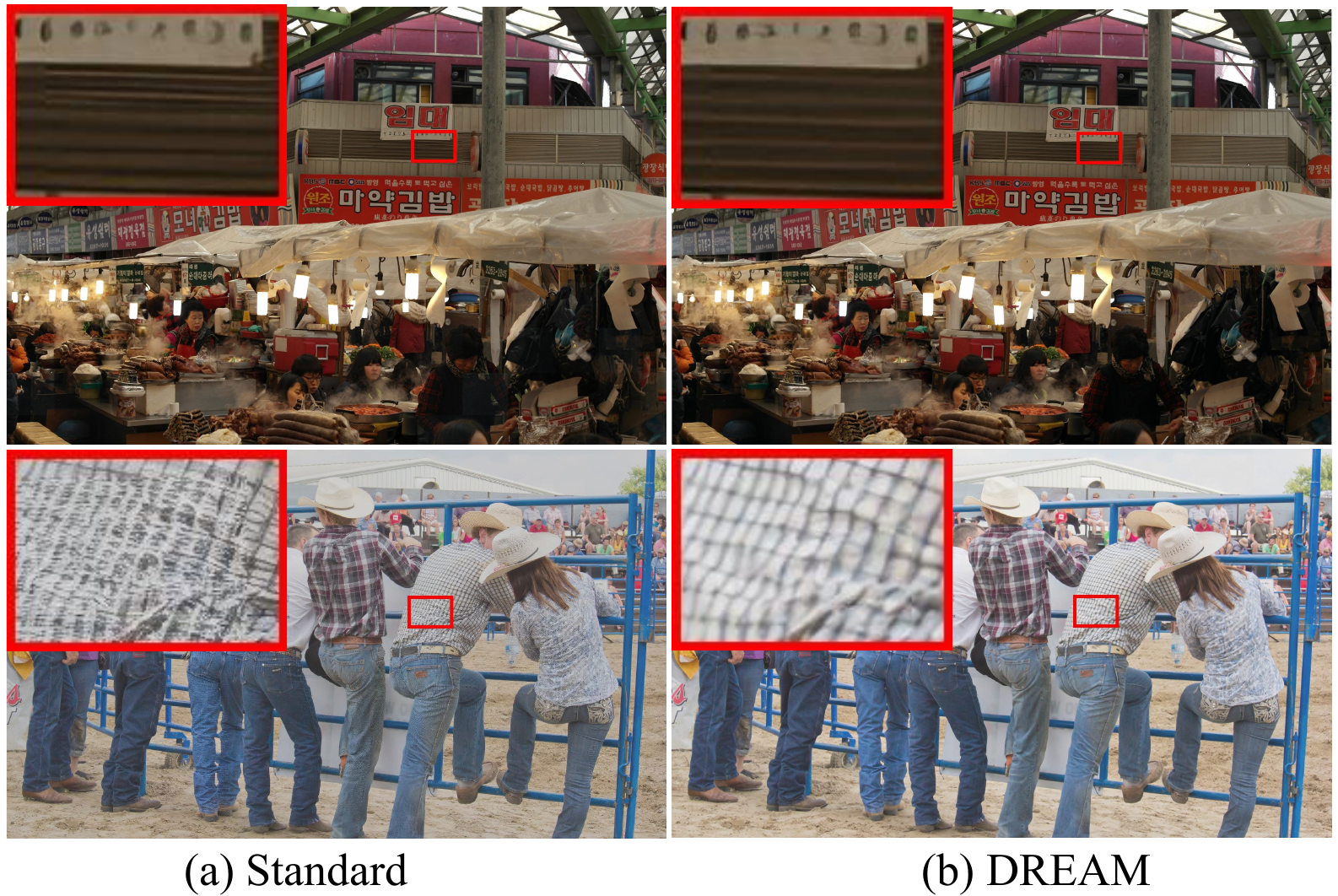}
    \vspace{-.1in}
    \caption{
    Qualitative comparison for $4\times$ SR on DIV2K~\cite{agustsson2017ntire}. Top with SR3~\cite{saharia2022image} the baseline; bottom with ResShift~\cite{yue2023resshift} the baseline.} 
        \label{fig:div}
    \vspace{-.1in}
\end{figure}

% \noindent
\textbf{General scene super-resolution.} \Cref{fig:div} shows a visual comparison of $4\times$ SR results on the DIV2K dataset~\cite{agustsson2017ntire}, using our DREAM approach against standard diffusion methods, with SR3 and ResShift as baselines. Standard training tends to produce images with blurred details and compromised realism, evident in unclear window outlines and distorted shirt textures. In contrast, DREAM maintains structural integrity and delivers more realistic textures. Following~\cite{guo2022lar}, we conduct a comprehensive comparison with various regression-based and generative methods on the DIV2K dataset. The results, detailed in~\Cref{tab:div2k} and benchmarked against models from~\cite{liang2021hierarchical}, demonstrate DREAM's effectiveness. Notably, DREAM has led to an increase of $1.08$dB and $3.14$dB in PSNR, and improvements of $0.03$ and $0.11$ in SSIM for SR3 and ResShift, respectively, outperforming other generative methods. Moreover, these methods demonstrate comparable or superior performance in perceptual quality metrics, marked by a 0.087 reduction in LPIPS for ResShift.  Although LPIPS scores are not as favorable as those obtained by HCFlow++, even with DREAM applied,  further improvements in image quality could be achieved through advanced network designs and incorporating GAN loss, as in HCFlow++.  However, such approaches are orthogonal to DREAM, and we leave these explorations for future work.

\begin{table}[]
% \label{tab:div2k}
\caption{Quantitative comparison for $4\times$ SR on DIV2K. All models are trained on DIV2K plus Flickr2K~\cite{timofte2017ntire}. The \colorbox{red!20}{best} and \colorbox{orange!20}{second-best} results among generative models are colorized.
% {\color{red}Red} and {\color{blue}blue} colors indicate the best and the second-best performance among generative models, respectively.
% The bold values indicate the best results among generative models.
}
\vspace{-.1in}
\setlength{\tabcolsep}{0.8mm}{
\small
\begin{tabular}{clccc}
\hline
\multicolumn{2}{c}{Method}                                               & PSNR$\uparrow$                     & SSIM$\uparrow$            & LPIPS$\downarrow$\\ \hline
&{Bicubic}                                              & 26.7                      & 0.77       & 0.409                                   \\ \hline
\multirow{2}{*}{Reg.-based}   & EDSR \cite{lim2017enhanced}             & 28.98                     & 0.83    & 0.270                                     \\
                              & RRDB \cite{wang2018esrgan}               & 29.44                     & 0.84          & 0.253                              \\ \hline\hline
\multirow{2}{*}{GAN-based}    & ESRGAN \cite{wang2018esrgan}             & 26.22                     & 0.75   &    0.124                                 \\
                              & RankSRGAN \cite{zhang2019ranksrgan}       & 26.55                     & 0.75        &0.128                               \\ \hline
\multirow{2}{*}{Flow-based}   & SRFlow \cite{lugmayr2020srflow}          & 27.09                     & \cellcolor{orange!20}0.76      & \cellcolor{orange!20}0.121                                 \\
                              & HCFlow \cite{liang2021hierarchical}      & 27.02    & \cellcolor{orange!20}0.76      & 0.124\\ \hline
Flow+GAN                      & HCFlow++ \cite{liang2021hierarchical}    & 26.61                     & 0.74      & \cellcolor{red!20}0.110                                 \\ \hline
\multirow{4}{*}{Diffusion} &  SR3~\cite{saharia2022image} & 27.02 & \cellcolor{orange!20}0.76 & \cellcolor{orange!20}0.121     \\
 &  SR3~\cite{saharia2022image}+DREAM & \cellcolor{orange!20}28.10 & \cellcolor{red!20}0.79 & \cellcolor{orange!20}0.121     \\
 & ResShift~\cite{yue2023resshift} & 25.30 & 0.68 & 0.211\\
 & ResShift~\cite{yue2023resshift}+DREAM & \cellcolor{red!20}28.44 & \cellcolor{red!20}0.79 & 0.124\\\hline
\end{tabular}
\label{tab:div2k}}
\vspace{-.1in}
\end{table}

\subsection{Training and sampling acceleration}

\begin{figure}[t]
     \centering
     \begin{subfigure}[b]{0.23\textwidth}
         \centering
         \includegraphics[width=\textwidth]{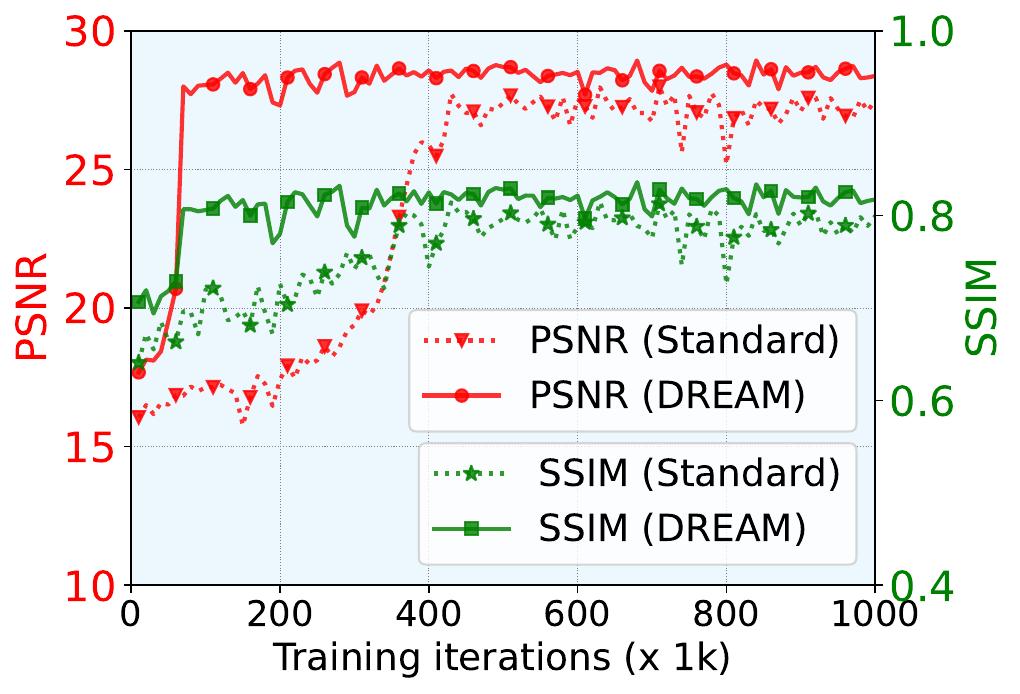}
         \caption{Distortion}
         \label{fig:training-ps}
     \end{subfigure}
     \begin{subfigure}[b]{0.242\textwidth}
         \centering
         \includegraphics[width=\textwidth]{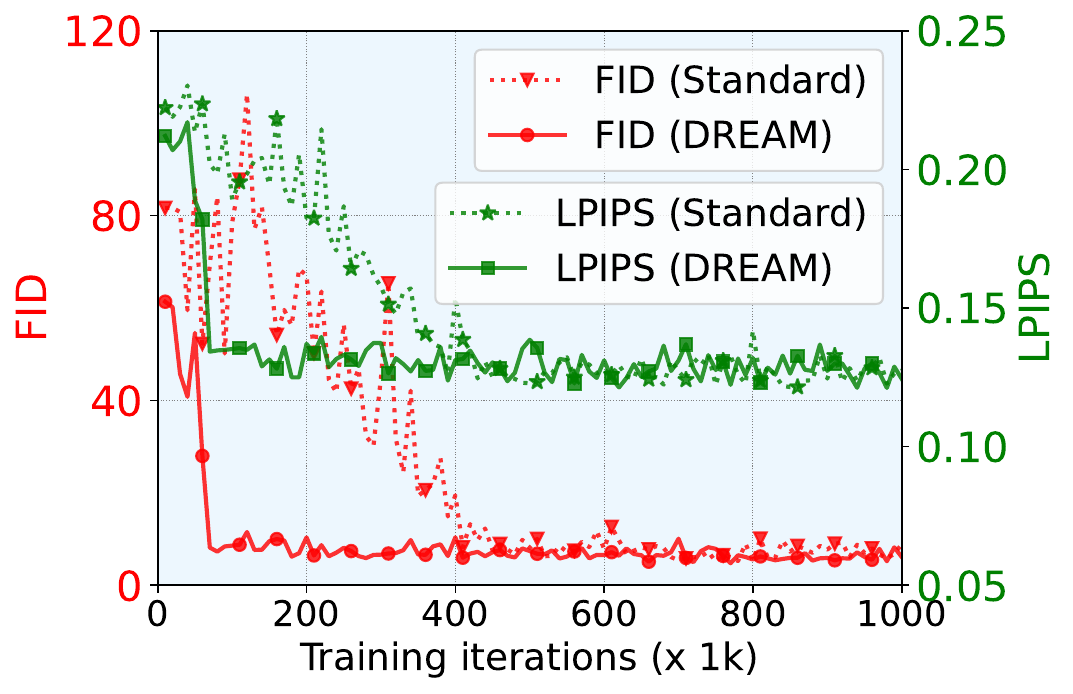}
         \caption{Perception}
         \label{fig:training-fl}
     \end{subfigure}
     \vspace{-.28in}
         \caption{Evolution of distortion metrics (left) and perceptual metrics (right) using SR3 as a baseline on the DIV2K dataset.}
        \label{fig:training}
        \vspace{-.1in}
\end{figure}

The DREAM strategy not only improves SR image quality but also accelerates the training.  As shown in~\Cref{fig:teaser}, DREAM reaches convergence at around 100k to 150k iterations, a significant improvement over the standard diffusion-based SR3's 400k iterations. Moreover, \Cref{fig:training} illustrates the  evolution of training in terms of distortion metrics (PSNR and SSIM) and perception metrics (LPIPS and FID) using SR3 as the baseline on the DIV2K dataset. DREAM not only converges faster but also surpasses SR3’s final results before its own convergence. For example, DREAM achieves a PSNR of $28.07$ and FID of $14.72$ at just $470$k iterations, while the baseline SR3 with standard diffusion reaches PSNR $27.02$ and FID $16.72$ after full convergence at $980$k iterations, indicating a \emph{$2\times$ speedup in training}.  Additional experiments with different baselines and datasets can be found in the appendix.
% More experiments using different baselines on various datasets are in the appendix. 

% \subsection{Sampling efficiency}

\begin{figure}[t]
     \centering
     \begin{subfigure}[b]{0.23\textwidth}
         \centering
         \includegraphics[width=\textwidth]{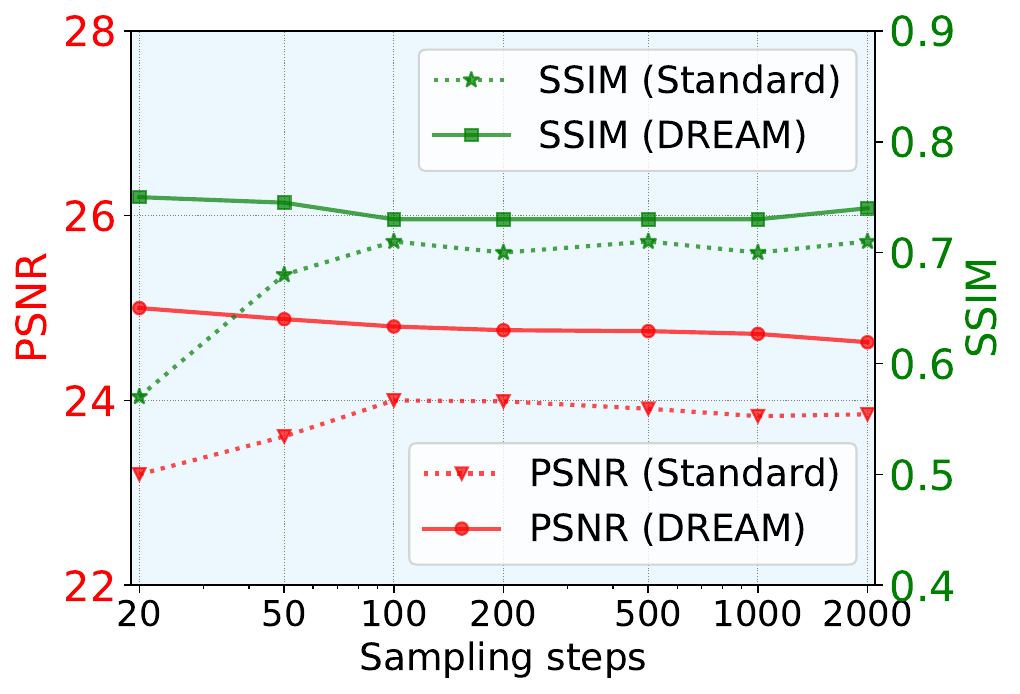}
         \caption{Distortion}
         \label{fig:sampling-ps}
     \end{subfigure}
     \begin{subfigure}[b]{0.242\textwidth}
         \centering
         \includegraphics[width=\textwidth]{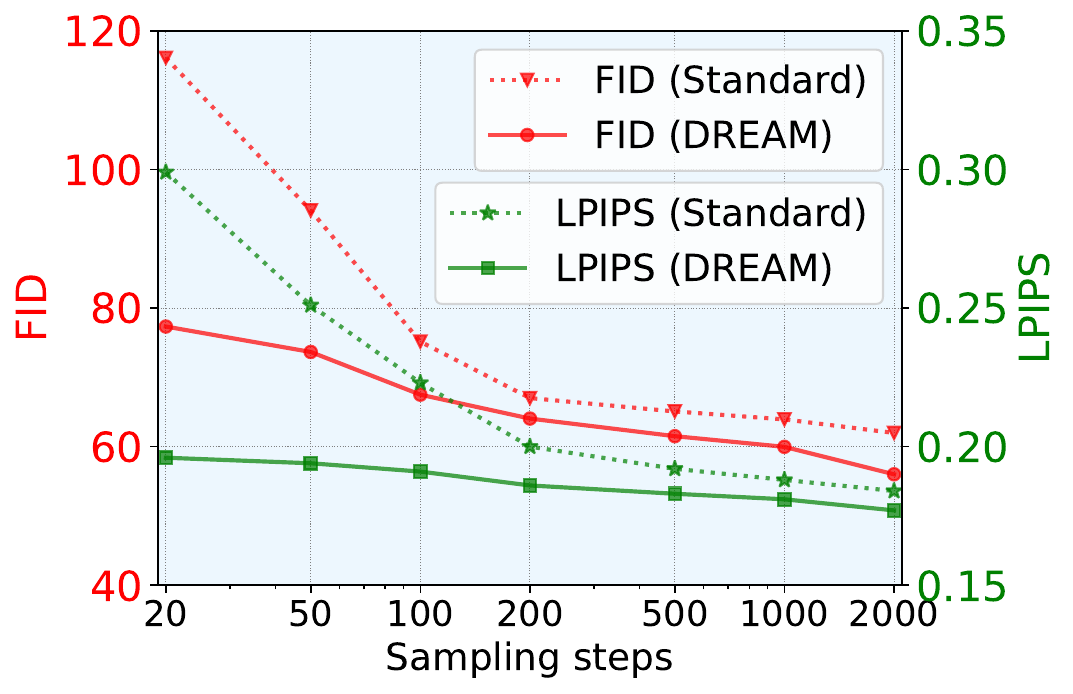}
         \caption{Perception}
         \label{fig:sampling-fl}
     \end{subfigure}
     \vspace{-.28in}
         \caption{Comparison of distortion metrics (left) and perception metrics (right) with varying sampling steps, using  SR3 as a baseline on the CelebA-HQ dataset.}
        \label{fig:sampling}
        \vspace{-.1in}
\end{figure}

Moreover, DREAM considerably accelerates the sampling process, outperforming standard diffusion training with fewer sampling steps.  \Cref{fig:sampling} demonstrates this using SR3 on the CelebA-HQ dataset, comparing SR images generated with varying sampling steps in terms of both distortion and perception metrics. While the standard baseline typically requires an entire 2000 sampling steps, DREAM achieves improved distortion metrics  ($0.73$ v.s. $0.71$ in SSIM) and comparable perceptual quality ($0.189$ v.s. $0.184$ in LPIPS) with only 100 steps. This marks \emph{$20\times$ speedup in sampling}. More details are available in the appendix.

\subsection{Out-of-distribution (OOD) evaluations}

\begin{figure}[t]
\centering
\includegraphics[width=0.45\textwidth]{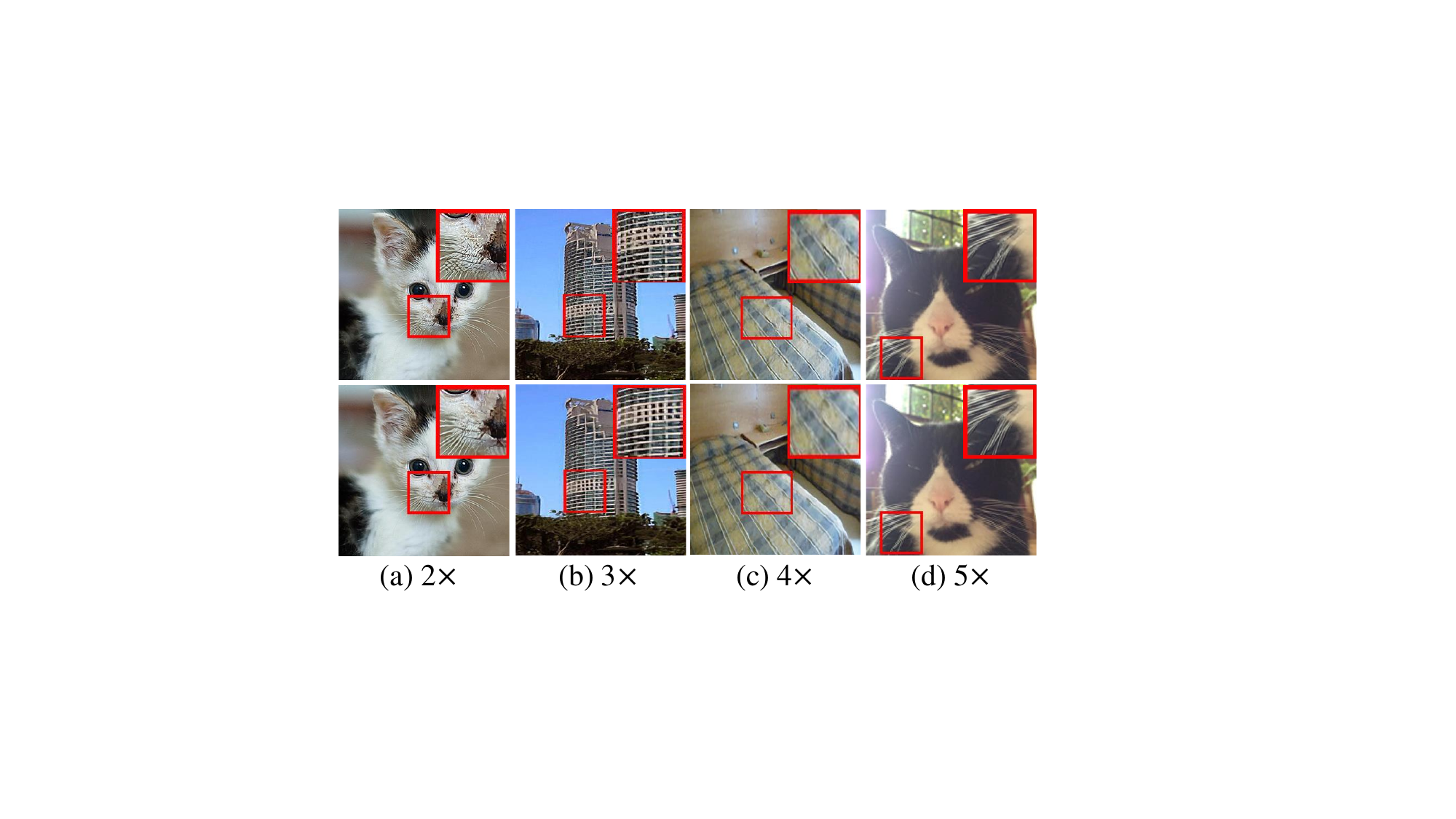} % Reduce the figure size so that it is slightly narrower than the column.
\vspace{-.1in}
\caption{Visual comparison of OOD SR. We use SR3 as a baseline, pretrain it on DIV2K and evaluate on CAT and LSUN, across various scales. The top row is obtained using standard training for SR3; the bottom row is generated using DREAM on SR3. }
\label{fig:lsun}
% \vspace{-.1in}
\end{figure}

To evaluate our approach's OOD performance, we train the SR3 model on DIV2K for $4\times$ SR scaling, then evaluate its performance on various natural image datasets from the CAT~\cite{zhang2008cat} and LSUN~\cite{yu2015lsun} benchmarks, covering multiple SR scales. This OOD evaluation encompasses both dataset diversity and scaling differences. As shown in~\Cref{fig:lsun}, our DREAM training approach significantly enhances model robustness, producing more realistic and clearer images across different scales. For instance, it captures finer details such as the beard of cats at $2\times$ and $5\times$ scales, the structural integrity of a tower at $3\times$ scale, and the intricate wrinkles on a bed at $4\times$ scale.  Following~\cite{gao2023implicit}, \Cref{tab:nature} presents the average PSNR and LPIPS metrics for 100 selected images from these validation datasets. Our findings show that the DREAM training framework consistently improves baseline model across diverse datasets and scales.

\begin{table}[t]\small
\caption{Quantitative comparison of OOD SR on CAT and LSUN Bedroom and Tower validation sets at various scales.}
\vspace{-.1in}
\centering
\label{tab:nature}
\setlength{\tabcolsep}{0.8mm}{
\begin{tabular}{ccccc}
\hline
Scale &  & Cats & Towers & Bedrooms \\\hline
\multirow{2}{*}{$2\times$} & Standard & 19.72/0.398 &  18.82/0.333& 20.20/0.314 \\
 & DREAM & \textbf{22.50}/\textbf{0.337} & \textbf{20.89}/\textbf{0.288} & \textbf{22.15}/\textbf{0.278} \\\hline
 \multirow{2}{*}{$3\times$} & Standard & 22.48/0.281 & 18.42/0.266 & 20.14/0.235 \\
 & DREAM & \textbf{23.90}/\textbf{0.265} & \textbf{19.35}/\textbf{0.252} & \textbf{20.65}/\textbf{0.231} \\\hline
\multirow{2}{*}{$4\times$} & Standard & 26.49/0.257 & 24.03/0.217 & 26.89/0.187 \\
 & DREAM & \textbf{27.19}/\textbf{0.246} & \textbf{24.94}/\textbf{0.212} & \textbf{27.53}/\textbf{0.183} \\\hline
\multirow{2}{*}{$5\times$} & Standard & 24.52/0.381 & 21.79/0.331 & 23.18/0.313 \\
 & DREAM & \textbf{24.58}/\textbf{0.373} & \textbf{21.84}/\textbf{0.324} & \textbf{23.19}/\textbf{0.310}
 \\\hline
\end{tabular}}
\vspace{-.1in}
\end{table}

\section{Conclusion}
This paper introduces DREAM, a novel training framework designed to address the training-sampling discrepancy in conditional diffusion models with minimal code modifications. DREAM comprises two key components: diffusion rectification and estimation adaptation. Diffusion rectification extends the existing training framework for diffusion models by aligning training more closely with sampling through self-estimation. Estimation adaptation optimizes the balance between accuracy and fidelity by adaptively incorporating ground-truth information. When applied to SISR tasks, DREAM effectively bridges the gap between training and sampling. Extensive experiments demonstrate that DREAM enhances distortion and perception metrics across various diffusion-based SR baselines. It also speeds up training, improves sampling efficiency, and achieves robust OOD performance across diverse datasets and scales. 

While DREAM is mainly utilized for SR in this work, its capabilities are applicable to a range of dense visual prediction tasks. Future research may investigate its use in both low-level vision tasks, such as inpainting and deblurring, and high-level vision tasks like semantic segmentation and depth estimation. 
% Additionally, exploring DREAM's application in both unconditional and conditional image generation presents an intriguing direction for future work.

% \newpage

% \newpage
{
    \small
    % \nocite{ding2021cdfi,ding2022sparsity,geng2022rstt}
    \bibliographystyle{ieeenat_fullname}
    \bibliography{main}

\begin{thebibliography}{69}
\providecommand{\natexlab}[1]{#1}
\providecommand{\url}[1]{\texttt{#1}}
\expandafter\ifx\csname urlstyle\endcsname\relax
  \providecommand{\doi}[1]{doi: #1}\else
  \providecommand{\doi}{doi: \begingroup \urlstyle{rm}\Url}\fi

\bibitem[Agustsson and Timofte(2017)]{agustsson2017ntire}
Eirikur Agustsson and Radu Timofte.
\newblock Ntire 2017 challenge on single image super-resolution: Dataset and study.
\newblock In \emph{Proceedings of the IEEE conference on computer vision and pattern recognition workshops}, pages 126--135, 2017.

\bibitem[Anwar and Barnes(2020)]{anwar2020densely}
Saeed Anwar and Nick Barnes.
\newblock Densely residual laplacian super-resolution.
\newblock \emph{IEEE Transactions on Pattern Analysis and Machine Intelligence}, 44\penalty0 (3):\penalty0 1192--1204, 2020.

\bibitem[Bevilacqua et~al.(2012)Bevilacqua, Roumy, Guillemot, and Alberi-Morel]{bevilacqua2012low}
Marco Bevilacqua, Aline Roumy, Christine Guillemot, and Marie~Line Alberi-Morel.
\newblock Low-complexity single-image super-resolution based on nonnegative neighbor embedding.
\newblock 2012.

\bibitem[Blau and Michaeli(2018)]{blau2018perception}
Yochai Blau and Tomer Michaeli.
\newblock The perception-distortion tradeoff.
\newblock In \emph{Proceedings of the IEEE conference on computer vision and pattern recognition}, pages 6228--6237, 2018.

\bibitem[Chan et~al.(2021)Chan, Wang, Xu, Gu, and Loy]{chan2021glean}
Kelvin~CK Chan, Xintao Wang, Xiangyu Xu, Jinwei Gu, and Chen~Change Loy.
\newblock Glean: Generative latent bank for large-factor image super-resolution.
\newblock In \emph{Proceedings of the IEEE/CVF conference on computer vision and pattern recognition}, pages 14245--14254, 2021.

\bibitem[Chen et~al.(2018)Chen, Tai, Liu, Shen, and Yang]{chen2018fsrnet}
Yu Chen, Ying Tai, Xiaoming Liu, Chunhua Shen, and Jian Yang.
\newblock Fsrnet: End-to-end learning face super-resolution with facial priors.
\newblock In \emph{Proceedings of the IEEE conference on computer vision and pattern recognition}, pages 2492--2501, 2018.

\bibitem[Chen et~al.(2021)Chen, Liu, and Wang]{chen2021learning}
Yinbo Chen, Sifei Liu, and Xiaolong Wang.
\newblock Learning continuous image representation with local implicit image function.
\newblock In \emph{Proceedings of the IEEE/CVF conference on computer vision and pattern recognition}, pages 8628--8638, 2021.

\bibitem[Dhariwal and Nichol(2021)]{dhariwal2021diffusion}
Prafulla Dhariwal and Alexander Nichol.
\newblock Diffusion models beat gans on image synthesis.
\newblock \emph{Advances in neural information processing systems}, 34:\penalty0 8780--8794, 2021.

\bibitem[Ding et~al.(2021)Ding, Liang, Zhu, and Zharkov]{ding2021cdfi}
Tianyu Ding, Luming Liang, Zhihui Zhu, and Ilya Zharkov.
\newblock Cdfi: Compression-driven network design for frame interpolation.
\newblock In \emph{Proceedings of the IEEE/CVF conference on computer vision and pattern recognition}, pages 8001--8011, 2021.

\bibitem[Ding et~al.(2022)Ding, Liang, Zhu, Chen, and Zharkov]{ding2022sparsity}
Tianyu Ding, Luming Liang, Zhihui Zhu, Tianyi Chen, and Ilya Zharkov.
\newblock Sparsity-guided network design for frame interpolation.
\newblock \emph{arXiv preprint arXiv:2209.04551}, 2022.

\bibitem[Dinh et~al.(2016)Dinh, Sohl-Dickstein, and Bengio]{dinh2016density}
Laurent Dinh, Jascha Sohl-Dickstein, and Samy Bengio.
\newblock Density estimation using real {NVP}.
\newblock \emph{arXiv:1605.08803}, 2016.

\bibitem[Dong et~al.(2014)Dong, Loy, He, and Tang]{dong2014learning}
Chao Dong, Chen~Change Loy, Kaiming He, and Xiaoou Tang.
\newblock Learning a deep convolutional network for image super-resolution.
\newblock In \emph{Computer Vision--ECCV 2014: 13th European Conference, Zurich, Switzerland, September 6-12, 2014, Proceedings, Part IV 13}, pages 184--199. Springer, 2014.

\bibitem[Everaert et~al.(2023)Everaert, Fitsios, Bocchio, Arpa, S{\"u}sstrunk, and Achanta]{everaert2023exploiting}
Martin~Nicolas Everaert, Athanasios Fitsios, Marco Bocchio, Sami Arpa, Sabine S{\"u}sstrunk, and Radhakrishna Achanta.
\newblock Exploiting the signal-leak bias in diffusion models.
\newblock \emph{arXiv preprint arXiv:2309.15842}, 2023.

\bibitem[Gao et~al.(2023)Gao, Liu, Zeng, Xu, Li, Luo, Liu, Zhen, and Zhang]{gao2023implicit}
Sicheng Gao, Xuhui Liu, Bohan Zeng, Sheng Xu, Yanjing Li, Xiaoyan Luo, Jianzhuang Liu, Xiantong Zhen, and Baochang Zhang.
\newblock Implicit diffusion models for continuous super-resolution.
\newblock In \emph{Proceedings of the IEEE/CVF Conference on Computer Vision and Pattern Recognition}, pages 10021--10030, 2023.

\bibitem[Geng et~al.(2022)Geng, Liang, Ding, and Zharkov]{geng2022rstt}
Zhicheng Geng, Luming Liang, Tianyu Ding, and Ilya Zharkov.
\newblock Rstt: Real-time spatial temporal transformer for space-time video super-resolution.
\newblock In \emph{Proceedings of the IEEE/CVF Conference on Computer Vision and Pattern Recognition}, pages 17441--17451, 2022.

\bibitem[Goodfellow et~al.(2014)Goodfellow, Pouget-Abadie, Mirza, Xu, Warde-Farley, Ozair, Courville, and Bengio]{goodfellow2014generative}
Ian~J Goodfellow, Jean Pouget-Abadie, Mehdi Mirza, Bing Xu, David Warde-Farley, Sherjil Ozair, Aaron Courville, and Yoshua Bengio.
\newblock {Generative Adversarial Networks}.
\newblock \emph{{NIPS}}, 2014.

\bibitem[Guo et~al.(2022)Guo, Zhang, Wu, Wang, Zhang, and Wang]{guo2022lar}
Baisong Guo, Xiaoyun Zhang, Haoning Wu, Yu Wang, Ya Zhang, and Yan-Feng Wang.
\newblock Lar-sr: A local autoregressive model for image super-resolution.
\newblock In \emph{Proceedings of the IEEE/CVF Conference on Computer Vision and Pattern Recognition}, pages 1909--1918, 2022.

\bibitem[Heusel et~al.(2017)Heusel, Ramsauer, Unterthiner, Nessler, and Hochreiter]{heusel2017gans}
Martin Heusel, Hubert Ramsauer, Thomas Unterthiner, Bernhard Nessler, and Sepp Hochreiter.
\newblock Gans trained by a two time-scale update rule converge to a local nash equilibrium.
\newblock \emph{Advances in neural information processing systems}, 30, 2017.

\bibitem[Ho et~al.(2020)Ho, Jain, and Abbeel]{ho2020denoising}
Jonathan Ho, Ajay Jain, and Pieter Abbeel.
\newblock Denoising diffusion probabilistic models.
\newblock \emph{Advances in neural information processing systems}, 33:\penalty0 6840--6851, 2020.

\bibitem[Ho et~al.(2022)Ho, Saharia, Chan, Fleet, Norouzi, and Salimans]{ho2022cascaded}
Jonathan Ho, Chitwan Saharia, William Chan, David~J Fleet, Mohammad Norouzi, and Tim Salimans.
\newblock Cascaded diffusion models for high fidelity image generation.
\newblock \emph{The Journal of Machine Learning Research}, 23\penalty0 (1):\penalty0 2249--2281, 2022.

\bibitem[Ji et~al.(2023)Ji, Chen, Xie, Hong, Liu, Liu, Lu, Li, and Luo]{ji2023ddp}
Yuanfeng Ji, Zhe Chen, Enze Xie, Lanqing Hong, Xihui Liu, Zhaoqiang Liu, Tong Lu, Zhenguo Li, and Ping Luo.
\newblock Ddp: Diffusion model for dense visual prediction.
\newblock \emph{arXiv preprint arXiv:2303.17559}, 2023.

\bibitem[Jo et~al.(2021)Jo, Oh, Vajda, and Kim]{jo2021tackling}
Younghyun Jo, Seoung~Wug Oh, Peter Vajda, and Seon~Joo Kim.
\newblock Tackling the ill-posedness of super-resolution through adaptive target generation.
\newblock In \emph{Proceedings of the IEEE/CVF Conference on Computer Vision and Pattern Recognition}, pages 16236--16245, 2021.

\bibitem[Karras et~al.(2017)Karras, Aila, Laine, and Lehtinen]{karras2017progressive}
Tero Karras, Timo Aila, Samuli Laine, and Jaakko Lehtinen.
\newblock Progressive growing of gans for improved quality, stability, and variation.
\newblock \emph{arXiv preprint arXiv:1710.10196}, 2017.

\bibitem[Karras et~al.(2018)Karras, Aila, Laine, and Lehtinen]{karras2018ProGAN}
Tero Karras, Timo Aila, Samuli Laine, and Jaakko Lehtinen.
\newblock Progressive growing of gans for improved quality, stability, and variation.
\newblock In \emph{{ICLR}}, 2018.

\bibitem[Karras et~al.(2019)Karras, Laine, and Aila]{karras2019style}
Tero Karras, Samuli Laine, and Timo Aila.
\newblock A style-based generator architecture for generative adversarial networks.
\newblock In \emph{Proceedings of the IEEE/CVF conference on computer vision and pattern recognition}, pages 4401--4410, 2019.

\bibitem[Kingma and Dhariwal(2018)]{Kingma2018}
Diederik~P. Kingma and Prafulla Dhariwal.
\newblock {Glow: Generative Flow with Invertible 1x1 Convolutions}.
\newblock In \emph{NIPS}, 2018.

\bibitem[Kingma and Welling(2013)]{Kingma2013}
Diederik~P Kingma and Max Welling.
\newblock {Auto-Encoding Variational Bayes}.
\newblock In \emph{ICLR}, 2013.

\bibitem[Ledig et~al.(2017)Ledig, Theis, Husz{\'a}r, Caballero, Cunningham, Acosta, Aitken, Tejani, Totz, Wang, et~al.]{ledig2017photo}
Christian Ledig, Lucas Theis, Ferenc Husz{\'a}r, Jose Caballero, Andrew Cunningham, Alejandro Acosta, Andrew Aitken, Alykhan Tejani, Johannes Totz, Zehan Wang, et~al.
\newblock Photo-realistic single image super-resolution using a generative adversarial network.
\newblock In \emph{Proceedings of the IEEE conference on computer vision and pattern recognition}, pages 4681--4690, 2017.

\bibitem[Li et~al.(2022)Li, Yang, Chang, Chen, Feng, Xu, Li, and Chen]{li2022srdiff}
Haoying Li, Yifan Yang, Meng Chang, Shiqi Chen, Huajun Feng, Zhihai Xu, Qi Li, and Yueting Chen.
\newblock Srdiff: Single image super-resolution with diffusion probabilistic models.
\newblock \emph{Neurocomputing}, 479:\penalty0 47--59, 2022.

\bibitem[Li et~al.(2023)Li, Qu, Sun, and Moens]{li2023alleviating}
Mingxiao Li, Tingyu Qu, Wei Sun, and Marie-Francine Moens.
\newblock Alleviating exposure bias in diffusion models through sampling with shifted time steps.
\newblock \emph{arXiv preprint arXiv:2305.15583}, 2023.

\bibitem[Liang et~al.(2021{\natexlab{a}})Liang, Cao, Sun, Zhang, Van~Gool, and Timofte]{liang2021swinir}
Jingyun Liang, Jiezhang Cao, Guolei Sun, Kai Zhang, Luc Van~Gool, and Radu Timofte.
\newblock Swinir: Image restoration using swin transformer.
\newblock In \emph{Proceedings of the IEEE/CVF international conference on computer vision}, pages 1833--1844, 2021{\natexlab{a}}.

\bibitem[Liang et~al.(2021{\natexlab{b}})Liang, Lugmayr, Zhang, Danelljan, Van~Gool, and Timofte]{liang2021hierarchical}
Jingyun Liang, Andreas Lugmayr, Kai Zhang, Martin Danelljan, Luc Van~Gool, and Radu Timofte.
\newblock Hierarchical conditional flow: A unified framework for image super-resolution and image rescaling.
\newblock In \emph{Proceedings of the IEEE/CVF International Conference on Computer Vision}, pages 4076--4085, 2021{\natexlab{b}}.

\bibitem[Liang et~al.(2022)Liang, Zeng, and Zhang]{liang2022details}
Jie Liang, Hui Zeng, and Lei Zhang.
\newblock Details or artifacts: A locally discriminative learning approach to realistic image super-resolution.
\newblock In \emph{Proceedings of the IEEE/CVF Conference on Computer Vision and Pattern Recognition}, pages 5657--5666, 2022.

\bibitem[Lim et~al.(2017)Lim, Son, Kim, Nah, and Mu~Lee]{lim2017enhanced}
Bee Lim, Sanghyun Son, Heewon Kim, Seungjun Nah, and Kyoung Mu~Lee.
\newblock Enhanced deep residual networks for single image super-resolution.
\newblock In \emph{Proceedings of the IEEE conference on computer vision and pattern recognition workshops}, pages 136--144, 2017.

\bibitem[Lugmayr et~al.(2020)Lugmayr, Danelljan, Van~Gool, and Timofte]{lugmayr2020srflow}
Andreas Lugmayr, Martin Danelljan, Luc Van~Gool, and Radu Timofte.
\newblock Srflow: Learning the super-resolution space with normalizing flow.
\newblock In \emph{Computer Vision--ECCV 2020: 16th European Conference, Glasgow, UK, August 23--28, 2020, Proceedings, Part V 16}, pages 715--732. Springer, 2020.

\bibitem[Menon et~al.(2020)Menon, Damian, Hu, Ravi, and Rudin]{menon2020pulse}
Sachit Menon, Alexandru Damian, Shijia Hu, Nikhil Ravi, and Cynthia Rudin.
\newblock Pulse: Self-supervised photo upsampling via latent space exploration of generative models.
\newblock In \emph{Proceedings of the ieee/cvf conference on computer vision and pattern recognition}, pages 2437--2445, 2020.

\bibitem[Nichol and Dhariwal(2021)]{nichol2021improved}
Alexander~Quinn Nichol and Prafulla Dhariwal.
\newblock Improved denoising diffusion probabilistic models.
\newblock In \emph{International Conference on Machine Learning}, pages 8162--8171. PMLR, 2021.

\bibitem[Ning et~al.(2023{\natexlab{a}})Ning, Li, Su, Salah, and Ertugrul]{ning2023elucidating}
Mang Ning, Mingxiao Li, Jianlin Su, Albert~Ali Salah, and Itir~Onal Ertugrul.
\newblock Elucidating the exposure bias in diffusion models.
\newblock \emph{arXiv preprint arXiv:2308.15321}, 2023{\natexlab{a}}.

\bibitem[Ning et~al.(2023{\natexlab{b}})Ning, Sangineto, Porrello, Calderara, and Cucchiara]{ning2023input}
Mang Ning, Enver Sangineto, Angelo Porrello, Simone Calderara, and Rita Cucchiara.
\newblock Input perturbation reduces exposure bias in diffusion models.
\newblock \emph{arXiv preprint arXiv:2301.11706}, 2023{\natexlab{b}}.

\bibitem[Oord et~al.(2016{\natexlab{a}})Oord, Dieleman, Zen, Simonyan, Vinyals, Graves, Kalchbrenner, Senior, and Kavukcuoglu]{oord-arxiv-2016}
A{\" a}ron van~den Oord, Sander Dieleman, Heiga Zen, Karen Simonyan, Oriol Vinyals, Alex Graves, Nal Kalchbrenner, Andrew Senior, and Koray Kavukcuoglu.
\newblock {WaveNet: A Generative Model for Raw Audio}.
\newblock \emph{arXiv preprint arXiv:1609.03499}, 2016{\natexlab{a}}.

\bibitem[Oord et~al.(2016{\natexlab{b}})Oord, Kalchbrenner, Vinyals, Espeholt, Graves, and Kavukcuoglu]{oord-nips-2016}
A{\" a}ron van~den Oord, Nal Kalchbrenner, Oriol Vinyals, Lasse Espeholt, Alex Graves, and Koray Kavukcuoglu.
\newblock {Conditional Image Generation with PixelCNN Decoders}.
\newblock In \emph{{NIPS}}, 2016{\natexlab{b}}.

\bibitem[Radford et~al.(2015)Radford, Metz, and Chintala]{radford2015unsupervised}
Alec Radford, Luke Metz, and Soumith Chintala.
\newblock Unsupervised representation learning with deep convolutional generative adversarial networks.
\newblock \emph{arXiv preprint arXiv:1511.06434}, 2015.

\bibitem[Rombach et~al.(2022)Rombach, Blattmann, Lorenz, Esser, and Ommer]{rombach2022high}
Robin Rombach, Andreas Blattmann, Dominik Lorenz, Patrick Esser, and Bj{\"o}rn Ommer.
\newblock High-resolution image synthesis with latent diffusion models.
\newblock In \emph{Proceedings of the IEEE/CVF conference on computer vision and pattern recognition}, pages 10684--10695, 2022.

\bibitem[Saharia et~al.(2022)Saharia, Ho, Chan, Salimans, Fleet, and Norouzi]{saharia2022image}
Chitwan Saharia, Jonathan Ho, William Chan, Tim Salimans, David~J Fleet, and Mohammad Norouzi.
\newblock Image super-resolution via iterative refinement.
\newblock \emph{IEEE Transactions on Pattern Analysis and Machine Intelligence}, 45\penalty0 (4):\penalty0 4713--4726, 2022.

\bibitem[Savinov et~al.(2022)Savinov, Chung, Binkowski, Elsen, and van~den Oord]{savinov2022stepunrolled}
Nikolay Savinov, Junyoung Chung, Mikolaj Binkowski, Erich Elsen, and Aaron van~den Oord.
\newblock Step-unrolled denoising autoencoders for text generation.
\newblock In \emph{International Conference on Learning Representations}, 2022.

\bibitem[Saxena et~al.(2023)Saxena, Kar, Norouzi, and Fleet]{saxena2023monocular}
Saurabh Saxena, Abhishek Kar, Mohammad Norouzi, and David~J Fleet.
\newblock Monocular depth estimation using diffusion models.
\newblock \emph{arXiv preprint arXiv:2302.14816}, 2023.

\bibitem[Soh et~al.(2019)Soh, Park, Jo, and Cho]{soh2019natural}
Jae~Woong Soh, Gu~Yong Park, Junho Jo, and Nam~Ik Cho.
\newblock Natural and realistic single image super-resolution with explicit natural manifold discrimination.
\newblock In \emph{Proceedings of the IEEE/CVF conference on computer vision and pattern recognition}, pages 8122--8131, 2019.

\bibitem[Sohl-Dickstein et~al.(2015)Sohl-Dickstein, Weiss, Maheswaranathan, and Ganguli]{sohl2015deep}
Jascha Sohl-Dickstein, Eric Weiss, Niru Maheswaranathan, and Surya Ganguli.
\newblock Deep unsupervised learning using nonequilibrium thermodynamics.
\newblock In \emph{International conference on machine learning}, pages 2256--2265. PMLR, 2015.

\bibitem[Song et~al.(2021)Song, Meng, and Ermon]{song2021denoising}
Jiaming Song, Chenlin Meng, and Stefano Ermon.
\newblock Denoising diffusion implicit models.
\newblock In \emph{International Conference on Learning Representations}, 2021.

\bibitem[Sun et~al.(2010)Sun, Xu, and Shum]{sun2010gradient}
Jian Sun, Zongben Xu, and Heung-Yeung Shum.
\newblock Gradient profile prior and its applications in image super-resolution and enhancement.
\newblock \emph{IEEE Transactions on Image Processing}, 20\penalty0 (6):\penalty0 1529--1542, 2010.

\bibitem[Szegedy et~al.(2016)Szegedy, Vanhoucke, Ioffe, Shlens, and Wojna]{szegedy2016rethinking}
Christian Szegedy, Vincent Vanhoucke, Sergey Ioffe, Jon Shlens, and Zbigniew Wojna.
\newblock Rethinking the inception architecture for computer vision.
\newblock In \emph{Proceedings of the IEEE conference on computer vision and pattern recognition}, pages 2818--2826, 2016.

\bibitem[Timofte et~al.(2017)Timofte, Agustsson, Van~Gool, Yang, and Zhang]{timofte2017ntire}
Radu Timofte, Eirikur Agustsson, Luc Van~Gool, Ming-Hsuan Yang, and Lei Zhang.
\newblock Ntire 2017 challenge on single image super-resolution: Methods and results.
\newblock In \emph{Proceedings of the IEEE conference on computer vision and pattern recognition workshops}, pages 114--125, 2017.

\bibitem[Vahdat and Kautz(2020)]{vahdat2021nvae}
Arash Vahdat and Jan Kautz.
\newblock {NVAE}: A deep hierarchical variational autoencoder.
\newblock In \emph{{NeurIPS}}, 2020.

\bibitem[Van Den~Oord et~al.(2016)Van Den~Oord, Kalchbrenner, and Kavukcuoglu]{van2016pixel}
A{\"a}ron Van Den~Oord, Nal Kalchbrenner, and Koray Kavukcuoglu.
\newblock Pixel recurrent neural networks.
\newblock In \emph{International conference on machine learning}, pages 1747--1756. PMLR, 2016.

\bibitem[Van Den~Oord et~al.(2017)Van Den~Oord, Vinyals, et~al.]{van2017neural}
Aaron Van Den~Oord, Oriol Vinyals, et~al.
\newblock Neural discrete representation learning.
\newblock \emph{Advances in neural information processing systems}, 30, 2017.

\bibitem[Wang et~al.(2018{\natexlab{a}})Wang, Yu, Dong, and Loy]{wang2018recovering}
Xintao Wang, Ke Yu, Chao Dong, and Chen~Change Loy.
\newblock Recovering realistic texture in image super-resolution by deep spatial feature transform.
\newblock In \emph{Proceedings of the IEEE conference on computer vision and pattern recognition}, pages 606--615, 2018{\natexlab{a}}.

\bibitem[Wang et~al.(2018{\natexlab{b}})Wang, Yu, Wu, Gu, Liu, Dong, Qiao, and Change~Loy]{wang2018esrgan}
Xintao Wang, Ke Yu, Shixiang Wu, Jinjin Gu, Yihao Liu, Chao Dong, Yu Qiao, and Chen Change~Loy.
\newblock Esrgan: Enhanced super-resolution generative adversarial networks.
\newblock In \emph{Proceedings of the European conference on computer vision (ECCV) workshops}, pages 0--0, 2018{\natexlab{b}}.

\bibitem[Wang et~al.(2004)Wang, Bovik, Sheikh, and Simoncelli]{wang2004image}
Zhou Wang, Alan~C Bovik, Hamid~R Sheikh, and Eero~P Simoncelli.
\newblock Image quality assessment: from error visibility to structural similarity.
\newblock \emph{IEEE transactions on image processing}, 13\penalty0 (4):\penalty0 600--612, 2004.

\bibitem[Yan et~al.(2015)Yan, Xu, Yang, and Nguyen]{yan2015single}
Qing Yan, Yi Xu, Xiaokang Yang, and Truong~Q Nguyen.
\newblock Single image superresolution based on gradient profile sharpness.
\newblock \emph{IEEE Transactions on Image Processing}, 24\penalty0 (10):\penalty0 3187--3202, 2015.

\bibitem[Yu et~al.(2015)Yu, Seff, Zhang, Song, Funkhouser, and Xiao]{yu2015lsun}
Fisher Yu, Ari Seff, Yinda Zhang, Shuran Song, Thomas Funkhouser, and Jianxiong Xiao.
\newblock Lsun: Construction of a large-scale image dataset using deep learning with humans in the loop.
\newblock \emph{arXiv preprint arXiv:1506.03365}, 2015.

\bibitem[Yu et~al.(2023)Yu, Shen, Huang, Zhou, Li, and Zhao]{yu2023debias}
Hu Yu, Li Shen, Jie Huang, Man Zhou, Hongsheng Li, and Feng Zhao.
\newblock Debias the training of diffusion models.
\newblock \emph{arXiv preprint arXiv:2310.08442}, 2023.

\bibitem[Yue et~al.(2023)Yue, Wang, and Loy]{yue2023resshift}
Zongsheng Yue, Jianyi Wang, and Chen~Change Loy.
\newblock Resshift: Efficient diffusion model for image super-resolution by residual shifting.
\newblock In \emph{Thirty-seventh Conference on Neural Information Processing Systems}, 2023.

\bibitem[Zhang et~al.(2020)Zhang, Gool, and Timofte]{zhang2020deep}
Kai Zhang, Luc~Van Gool, and Radu Timofte.
\newblock Deep unfolding network for image super-resolution.
\newblock In \emph{Proceedings of the IEEE/CVF conference on computer vision and pattern recognition}, pages 3217--3226, 2020.

\bibitem[Zhang et~al.(2021)Zhang, Liang, Van~Gool, and Timofte]{zhang2021designing}
Kai Zhang, Jingyun Liang, Luc Van~Gool, and Radu Timofte.
\newblock Designing a practical degradation model for deep blind image super-resolution.
\newblock In \emph{Proceedings of the IEEE/CVF International Conference on Computer Vision}, pages 4791--4800, 2021.

\bibitem[Zhang et~al.(2018{\natexlab{a}})Zhang, Isola, Efros, Shechtman, and Wang]{zhang2018unreasonable}
Richard Zhang, Phillip Isola, Alexei~A Efros, Eli Shechtman, and Oliver Wang.
\newblock The unreasonable effectiveness of deep features as a perceptual metric.
\newblock In \emph{Proceedings of the IEEE conference on computer vision and pattern recognition}, pages 586--595, 2018{\natexlab{a}}.

\bibitem[Zhang et~al.(2008)Zhang, Sun, and Tang]{zhang2008cat}
Weiwei Zhang, Jian Sun, and Xiaoou Tang.
\newblock Cat head detection-how to effectively exploit shape and texture features.
\newblock In \emph{Computer Vision--ECCV 2008: 10th European Conference on Computer Vision, Marseille, France, October 12-18, 2008, Proceedings, Part IV 10}, pages 802--816. Springer, 2008.

\bibitem[Zhang et~al.(2019)Zhang, Liu, Dong, and Qiao]{zhang2019ranksrgan}
Wenlong Zhang, Yihao Liu, Chao Dong, and Yu Qiao.
\newblock Ranksrgan: Generative adversarial networks with ranker for image super-resolution.
\newblock In \emph{Proceedings of the IEEE/CVF International Conference on Computer Vision}, pages 3096--3105, 2019.

\bibitem[Zhang et~al.(2018{\natexlab{b}})Zhang, Li, Li, Wang, Zhong, and Fu]{zhang2018image}
Yulun Zhang, Kunpeng Li, Kai Li, Lichen Wang, Bineng Zhong, and Yun Fu.
\newblock Image super-resolution using very deep residual channel attention networks.
\newblock In \emph{Proceedings of the European conference on computer vision (ECCV)}, pages 286--301, 2018{\natexlab{b}}.

\bibitem[Zhang et~al.(2018{\natexlab{c}})Zhang, Tian, Kong, Zhong, and Fu]{zhang2018residual}
Yulun Zhang, Yapeng Tian, Yu Kong, Bineng Zhong, and Yun Fu.
\newblock Residual dense network for image super-resolution.
\newblock In \emph{Proceedings of the IEEE conference on computer vision and pattern recognition}, pages 2472--2481, 2018{\natexlab{c}}.

\end{thebibliography}
}

\clearpage
\setcounter{page}{1}
\maketitlesupplementary
In this supplementary material, we begin by describing more details of the evaluation metrics and experiment setup in \Cref{sec:app-setup}. In following \Cref{sec:app-exp}, we present more quantitative comparisons and visualization results on various baselines and datasets, which further demonstrates the effectiveness of our DREAM strategy. We conclude with a discussion of the ethical implications in \Cref{sec:app-ethic}. 

\section{Metrics and setups}
We provide a more comprehensive explanation of the metrics and the experiment settings employed in the main text of the paper.
\label{sec:app-setup}
\subsection{Metrics}
In this section, we will detail the metrics applied to measure image distortion and perception quality. The distortion metrics encompass Peak Signal-to-Noise Ratio (PSNR) and Structural Similarity Index Measure (SSIM), as well as Consistency the the perception measurement include the Learned Perceptual Image Patch Similarity (LPIPS) and the Fréchet Inception Distance (FID).

\textbf{Peak Signal-to-Noise Ratio (PSNR).} PSNR is an indicator of image reconstruction quality. However, its value in decibels (dB) presents certain constraints when assessing super-resolution tasks \cite{menon2020pulse}. Thus, it acts merely as a referential metric of image quality, comparing the maximum possible signal to the level of background noise. Generally, a higher PSNR suggests a lower degree of image distortion.

\textbf{Structure Similarity Index Measure (SSIM).} Building on the image distortion modeling framework~\cite{wang2004image}, the SSIM applies the principles of structural similarity, mirroring the functionality of the human visual system. It is adept at detecting local structural alterations within an image. SSIM measures image attributes such as luminance, contrast, and structure by employing the mean for luminance assessment, variance for contrast evaluation, and covariance to gauge structural integrity.

\textbf{Consistency.} Consistency is measured by calculating the MSE ($\times10^{-5}$) between the low-resolution inputs and their corresponding downsampled super-resolution outputs.

\textbf{Learned Perceptual Image Patch Similarity (LPIPS).} LPIPS evaluates the perceptual resemblance between generated images and their authentic counterparts by analyzing deep feature representations.

\textbf{Fréchet Inception Distance score (FID).} FID~\cite{heusel2017gans} assesses image quality by emulating human judgment of image resemblance. This is achieved by utilizing a pre-trained Inception-V3 network~\cite{szegedy2016rethinking} to contrast the distribution patterns of the generated images against the distributions of the original, ground-truth images.
\subsection{Setups}

In this section, we will provide detailed descriptions of the configurations for various baseline models as well as the datasets utilized in our experiments. 

\textbf{SR3 model on face dataset.} We train the SR3~\cite{saharia2022image} model on an upscaled $8\times$ FFHQ dataset for 1M iterations and evaluate on 100 images from the CelebA~\cite{karras2017progressive} validation dataset. During training, the LR images are consistently resized to $16\times16$ pixels, while the HR counterparts are scaled to $128\times128$ pixels. For the SR image generation, the LR images are first upscaled to $128\times128$ pixels using bicubic interpolation and serve as the conditioning input. In alignment with the DDPM~\cite{ho2020denoising},  the Adam optimizer is utilized with a fixed learning rate of $1e-4$ through the training phase. The training employs a batch size of $4$, incorporates a dropout rate of $0.2$, and utilizes a linear beta scheduler over $2000$ steps with a starting value of 
$10^{-6}$ and a final value of $10^{-2}$. A single
24GB NVIDIA RTX A5000 GPU is used under this situation.

\textbf{IDM model on face dataset.} Adhering to the offical implementation of  the IDM~\cite{gao2023implicit}, the model is trained on a $8\times$ FFHQ dataset for 1M iterations and evaluated on 100 images from the CelebA~\cite{karras2017progressive} validation dataset. Specifically, throughout training, LR images are consistently resized to $16\times16$ pixels, while their HR counterparts are scaled to $128\times128$ pixels. These LR images are then processed through a specialized LR conditioning network, which is stacked with a series of convolutional layers, bilinear downsampling filtering, and leaky ReLU activation to extract a hierarchy of multi-resolution features. These features are then employed as the conditioning input for the denoising network. The training employs the Adam optimizer with a constant learning rate of $10^{-4}$, a batch size of 32, and a dropout rate of 0.2. We implement a linear beta scheduler that advances over 2000 steps, starting from $10^{-6}$ and escalating to $10^{-2}$. This setup is supported by two 24GB NVIDIA RTX A5000.

\textbf{SR3 model on general scene dataset.} We train the SR3~\cite{saharia2022image} model on upscaled $4\times$ the training dataset comparising DIV2K~\cite{agustsson2017ntire} and Flicker2K~\cite{timofte2017ntire} for 1M iterations. Consistent with the SRDiff~\cite{li2022srdiff}, each image is cropped into patches 
of $160 \times 160$ as the HR ground truths. To produce the corresponding LR image patches of $40\times40$ pixels, the HR image patches are downscaled using a bicubic kernel. These LR image patches are then resized back to the HR dimensions using bicubic interpolation and are used as the conditioning input for the super-resolution process. For evaluation, the entire DIV2K validation set, consisting of 100 images, is utilized. The HR images are downsampled using a bicubic kernel to generate LR images, which are then cropped into $40\times40$ pixel patches with a 5-pixel overlap between adjacent patches. The SR3 model is applied to these LR patches to yield the SR predictions which are subsequently merged to form the final SR images. The model's training utilizes the Adam optimizer with a steady learning rate of $10^{-4}$, a batch size of 32 patches, and a dropout rate of 0.2. A linear beta scheduler is applied over 1000 steps, initiating at $10^{-6}$ and culminating at $10^{-2}$. This configuration is executed on two 24GB NVIDIA RTX A5000 GPUs.
% \begin{figure} [t]
%      \centering
%     \includegraphics[width=0.4\textwidth]{figures/training_vs_sampling_with_drm.pdf}
%      \caption{Evaluation of training-sampling discrepancy under our DRM framework. The mean curve over 100 samples at each time step $t$ is plotted, with the shaded area representing the standard deviation of each metric.  Here, $T=2000$.}
%     \label{fig:error-dynamic-drm}
%     \vspace{-.1in}
% \end{figure}

\textbf{ResShift on general scene datatset.} Training the ResShift model~\cite{yue2023resshift}uses a $4\times$ dataset, combining the training sets from DIV2K~\cite{agustsson2017ntire} and Flickr2K~\cite{timofte2017ntire} over 0.5M iterations. Similar as data process in the previous SR3 setting, each image is partitioned into patches of 256x256 pixels to serve as HR ground truths. The LR image patches, resized to 64x64 pixels, are derived by downscaling the HR patches with a bicubic kernel. The VQGAN encoder, pre-trained on the ImageNet dataset, processes these LR patches to distill salient features, furnishing the necessary conditioning input for the following latent denoiser network. For performance evaluation, we use the entire DIV2K validation set, which comprises 100 images. The HR images are downsampled to LR with a bicubic kernel, and then segmented into 64x64 pixel patches, maintaining an 8-pixel overlap between adjacent patches. The latent denoiser model is applied to the LR patches to generate the corresponding SR latent codes. These latent codes are subsequently processed by the VQGAN decoder to reconstruct the SR patches, thereby producing the final high-resolution super-resolution images. The training regimen employs the Adam optimizer with a consistent learning rate of $5\times10^{-5}$ and a batch size of 32 patches. A linear beta scheduler is utilized over 50 steps, selected evenly from a linearly spaced 2000-steps schedule beginning at $10^{-6}$ and increasing to $10^{-2}$. The training is conducted using two 24GB NVIDIA RTX A5000.

\begin{table}[t]
\vspace{-3em}
\caption{Comparison of training speed and memory usage. The values denote the ratio of DREAM$/$standard.}
\vspace{-.1in}
\centering
\label{tab:training-memory}
\setlength{\tabcolsep}{1.6mm}{
\begin{tabular}{c|cc|cc}
\toprule
& \multicolumn{2}{c|}{Face} & \multicolumn{2}{c}{DIV2K} \\
 & SR3 & IDM & SR3 & ResShift \\\midrule
 Training time& 1.38 & 1.21 & 1.24 & 1.08\\
Training memory & 1.06 & 1.11 & 1.09 & 1.13\\
\bottomrule
\end{tabular}}
\vspace{-1.5em}
\end{table}

\section{Additional experimental results} 
In this section, we begin by providing additional results on the acceleration of training and sampling across various baselines and datasets in \Cref{sec:app-efficiency}. Lastly, in \Cref{sec:app-vis}, we offer a more comprehensive visual comparison on the general scene dataset, using the SR3~\cite{saharia2022image} and ResShift~\cite{yue2023resshift} models as baselines. 
\label{sec:app-exp}
% \subsection{Further analysis of DRM}\label{sec:app-drm-exp}

\begin{figure}[t]
     \centering
     \begin{subfigure}[b]{0.23\textwidth}
         \centering
         \includegraphics[width=\textwidth]{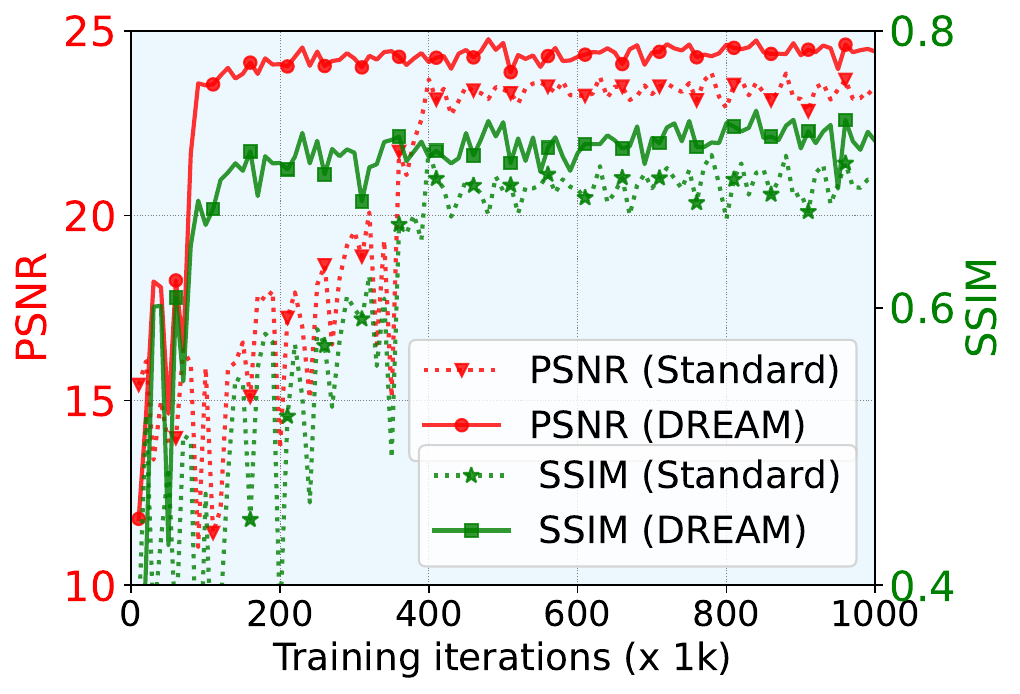}
         \caption{Distortion}
         \label{fig:sr3-face-training-ps}
     \end{subfigure}
     \begin{subfigure}[b]{0.242\textwidth}
         \centering
         \includegraphics[width=\textwidth]{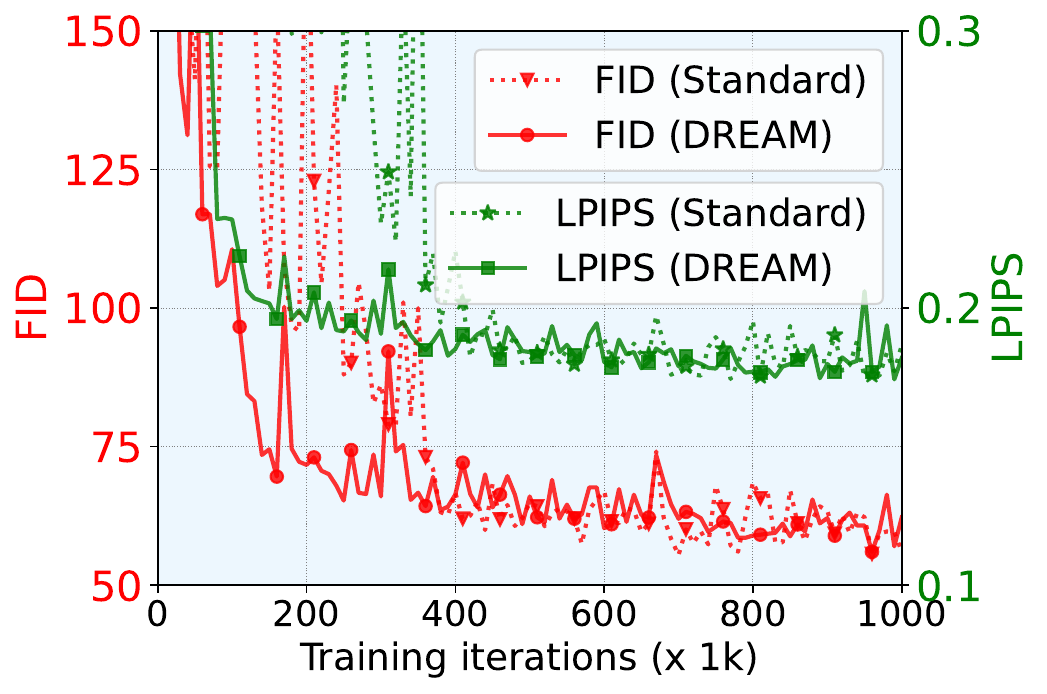}
         \caption{Perception}
         \label{fig:sr3-face-training-fl}
     \end{subfigure}
     \vspace{-.28in}
         \caption{Evolution of distortion metrics (left) and perceptual metrics (right) using SR3 as a baseline on the face dataset.}
        \label{fig:sr3-face-training}
        \vspace{-.1in}
\end{figure}

\begin{figure}[t]
     \centering
     \begin{subfigure}[b]{0.23\textwidth}
         \centering
         \includegraphics[width=\textwidth]{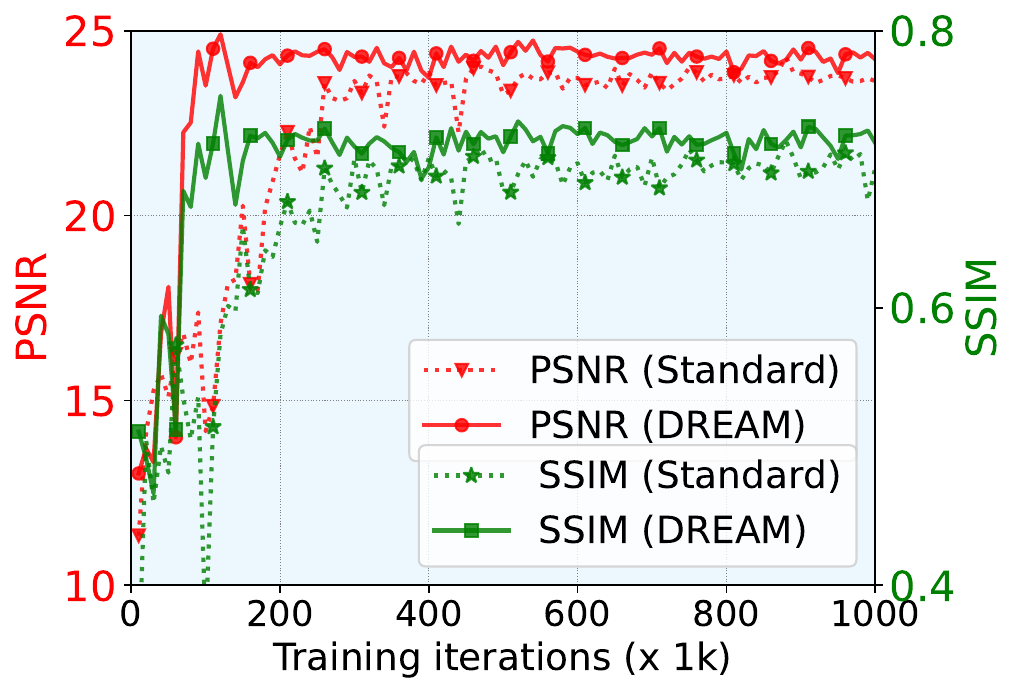}
         \caption{Distortion}
         \label{fig:idm-face-training-ps}
     \end{subfigure}
     \begin{subfigure}[b]{0.242\textwidth}
         \centering
         \includegraphics[width=\textwidth]{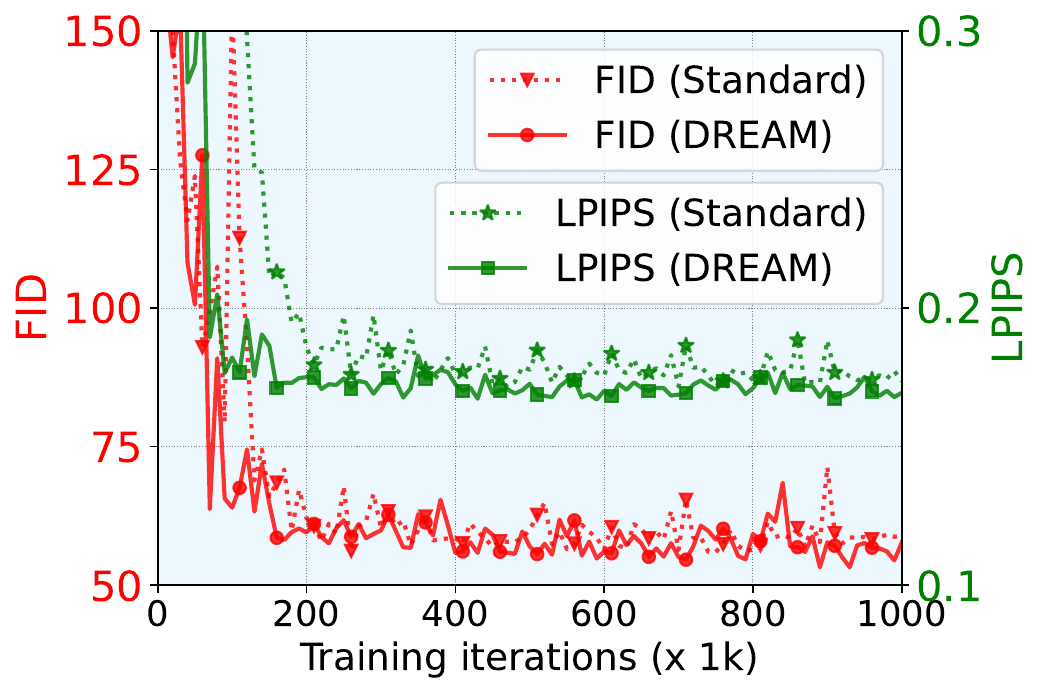}
         \caption{Perception}
         \label{fig:idm-face-training-fl}
     \end{subfigure}
     \vspace{-.28in}
         \caption{Evolution of distortion metrics (left) and perceptual metrics (right) using IDM as a baseline on the face dataset.}
        \label{fig:idm-face-training}
        \vspace{-.1in}
\end{figure}

\begin{figure}[h]
     \centering
     \begin{subfigure}[b]{0.23\textwidth}
         \centering
         \includegraphics[width=\textwidth]{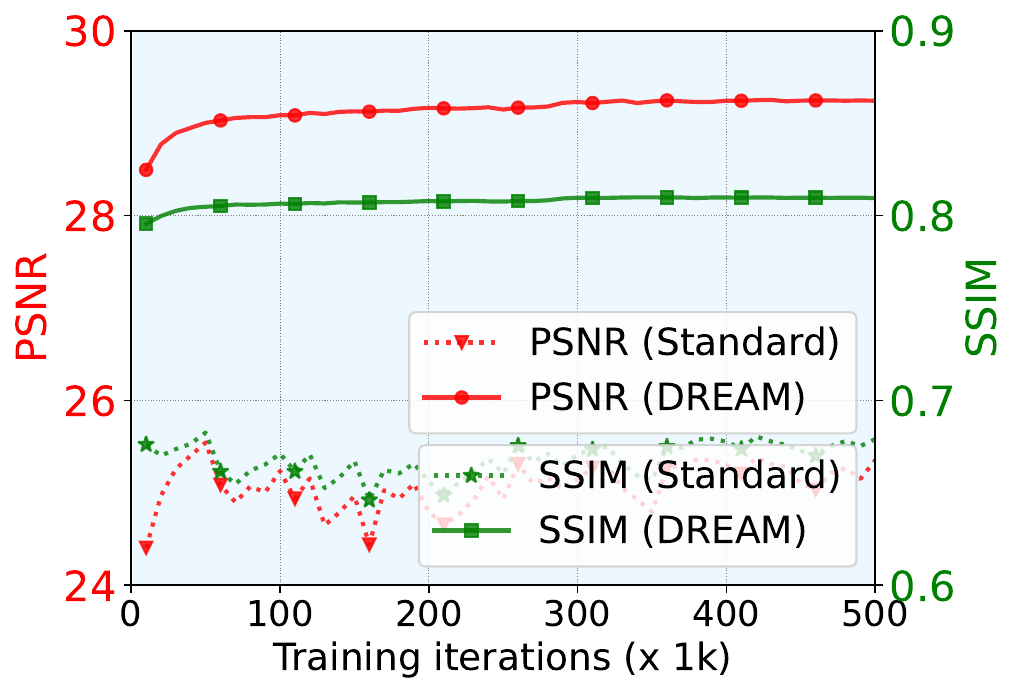}
         \caption{Distortion}
         \label{fig:res-div-training-ps}
     \end{subfigure}
     \begin{subfigure}[b]{0.242\textwidth}
         \centering
         \includegraphics[width=\textwidth]{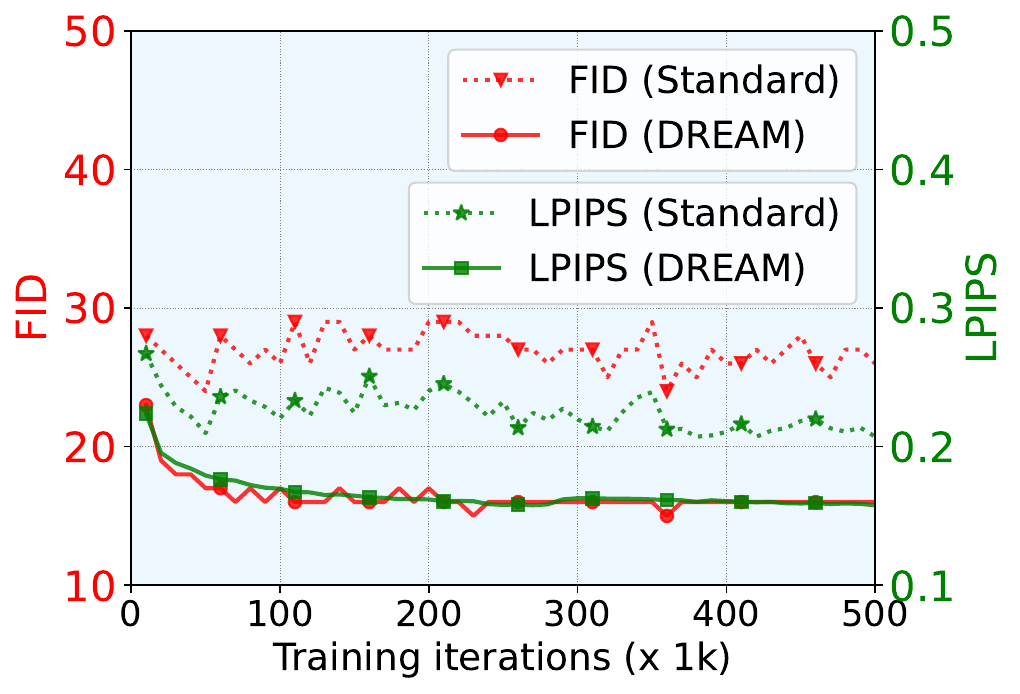}
         \caption{Perception}
         \label{fig:res-div-training-fl}
     \end{subfigure}
     \vspace{-.28in}
         \caption{Evolution of distortion metrics (left) and perceptual metrics (right) using ResShift as a baseline on the DIV2K dataset.}
        \label{fig:res-div-training}
        \vspace{-.1in}
\end{figure}

\subsection{Training and sampling acceleration}\label{sec:app-efficiency}

\textbf{Training efficiency.} In \Cref{tab:training-memory}, we detail the relative ratio of training speed and memory usage between our DREAM methodology and standard training approaches across a variety of baselines and datasets. Our DREAM method, which includes only a single additional forward computation, results in a marginal increase in training time (around $1.1\sim1.4\times$) and memory usage (approximately $1.05\sim1.15\times$). However, it offers a considerable advantage by significantly accelerating training convergence. We further illustrate the evolution of training through distortion metrics, namely PSNR and SSIM, as well as perception metrics such as LPIPS and FID. Utilizing SR3 and IDM as baselines for the face dataset, the improvements are evident in \Cref{fig:sr3-face-training} and \Cref{fig:idm-face-training}. The ResShift model, used as a baseline for the DIV2K dataset, demonstrates similar enhancements in \Cref{fig:res-div-training}. Notably, DREAM not only facilitates quicker convergence but also outperforms the final outcomes of several baselines after they fully converge. For example, with the face dataset, the SR3 model using DREAM achieves a PSNR of 24.49 and an FID of 61.02 in just 490k iterations, whereas the standard diffusion baseline reaches a PSNR of 23.85 and an FID of 61.98 after 880k iterations. This underlines a substantial training speedup by roughly $2 \times$ with DREAM. Similarly, the IDM model with DREAM reaches a PSNR of 23.54 and an FID of 55.81 in only 330k iterations, compared to the baseline achieving a PSNR of 23.85 and an FID of 61.98 after 760k iterations, reinforcing the significant efficiency of DREAM.

\begin{figure}[t]
     \centering
     \begin{subfigure}[b]{0.23\textwidth}
         \centering
         \includegraphics[width=\textwidth]{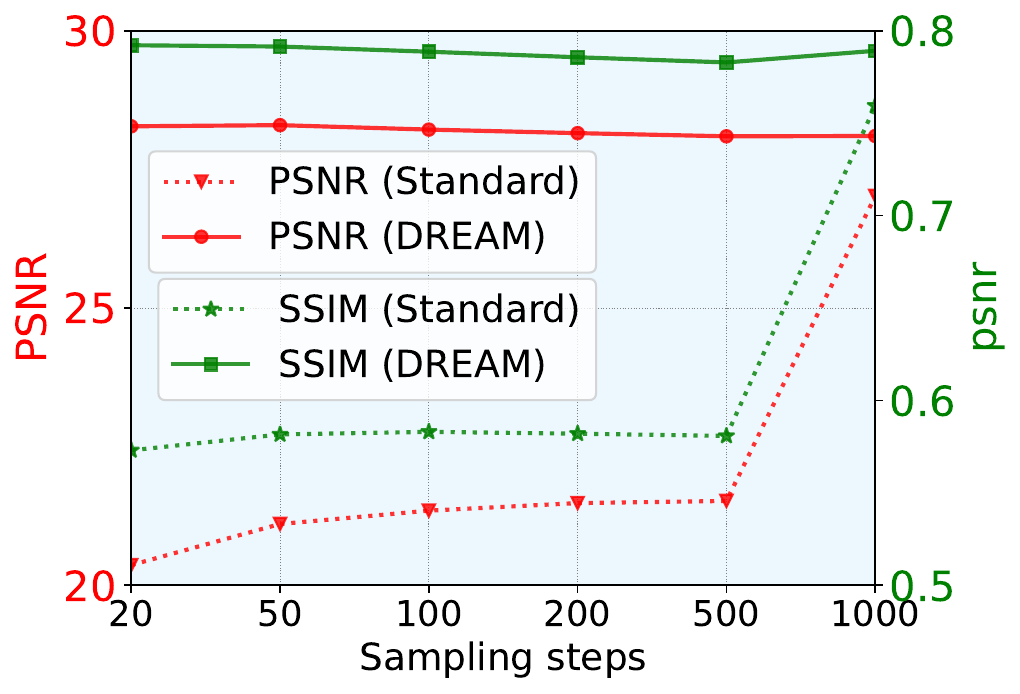}
         \caption{Distortion}
         \label{fig:idm-face-sampling-ps}
     \end{subfigure}
     \begin{subfigure}[b]{0.242\textwidth}
         \centering
         \includegraphics[width=\textwidth]{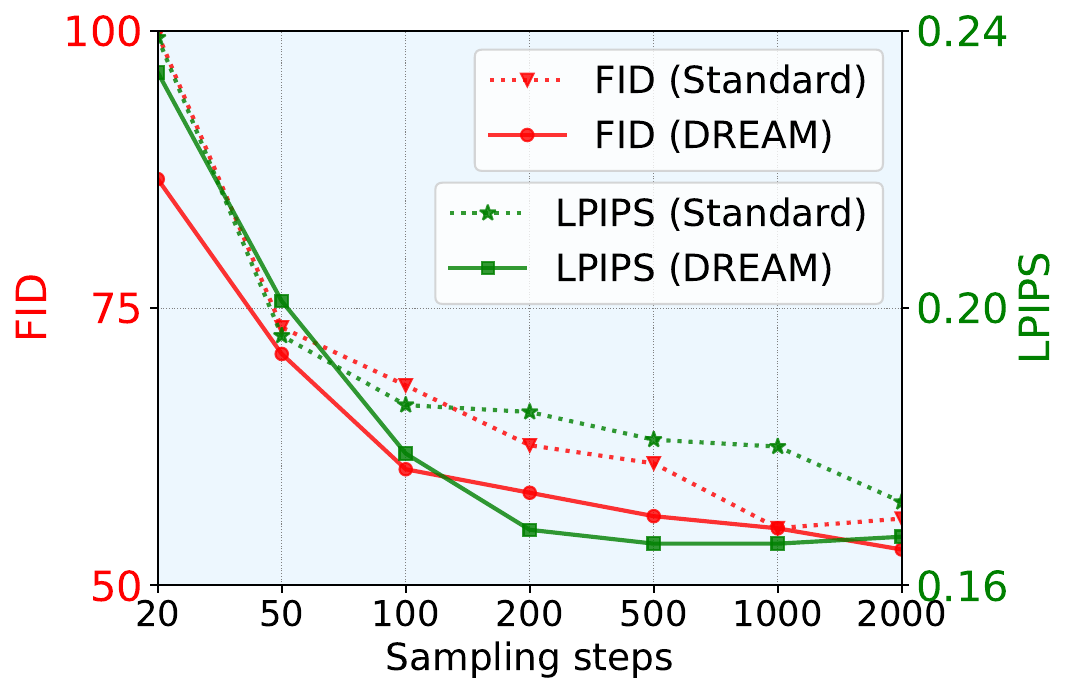}
         \caption{Perception}
         \label{fig:idm-face-sampling-fl}
     \end{subfigure}
     \vspace{-.28in}
         \caption{Comparison of distortion metrics (left) and perception metrics (right) with varying sampling steps, using  IDM as a baseline on the CelebA-HQ dataset.}
        \label{fig:idm-face-sampling}
        \vspace{-.2in}
\end{figure}

\begin{figure}[t]
     \centering
     \begin{subfigure}[b]{0.23\textwidth}
         \centering
         \includegraphics[width=\textwidth]{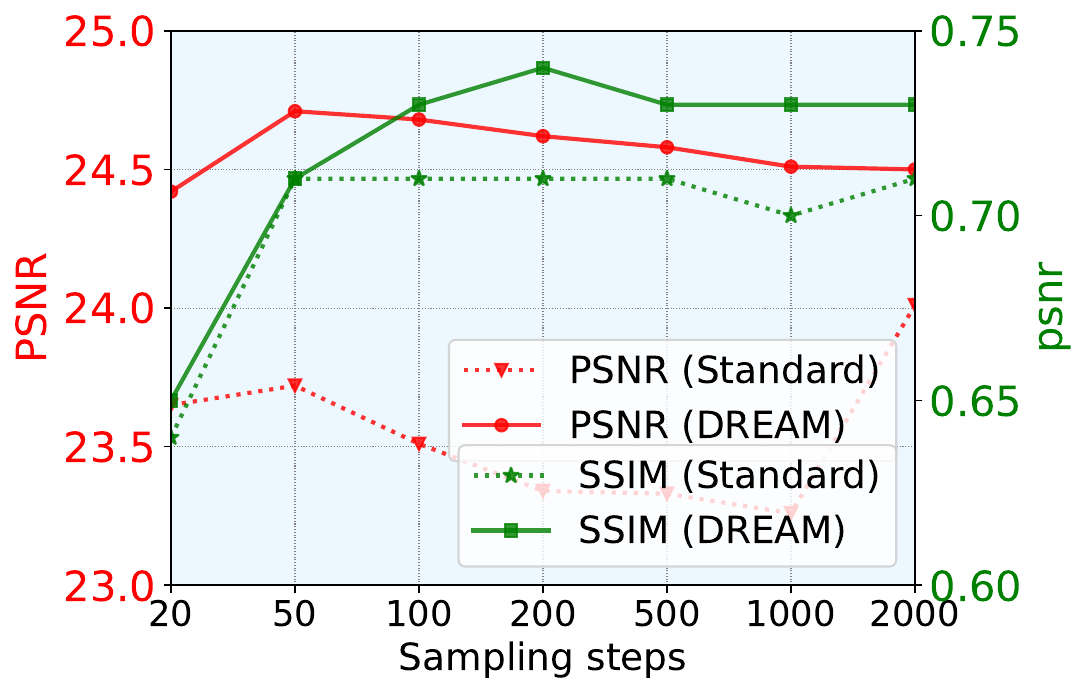}
         \caption{Distortion}
         \label{fig:sr3-div-sampling-ps}
     \end{subfigure}
     \begin{subfigure}[b]{0.242\textwidth}
         \centering
         \includegraphics[width=\textwidth]{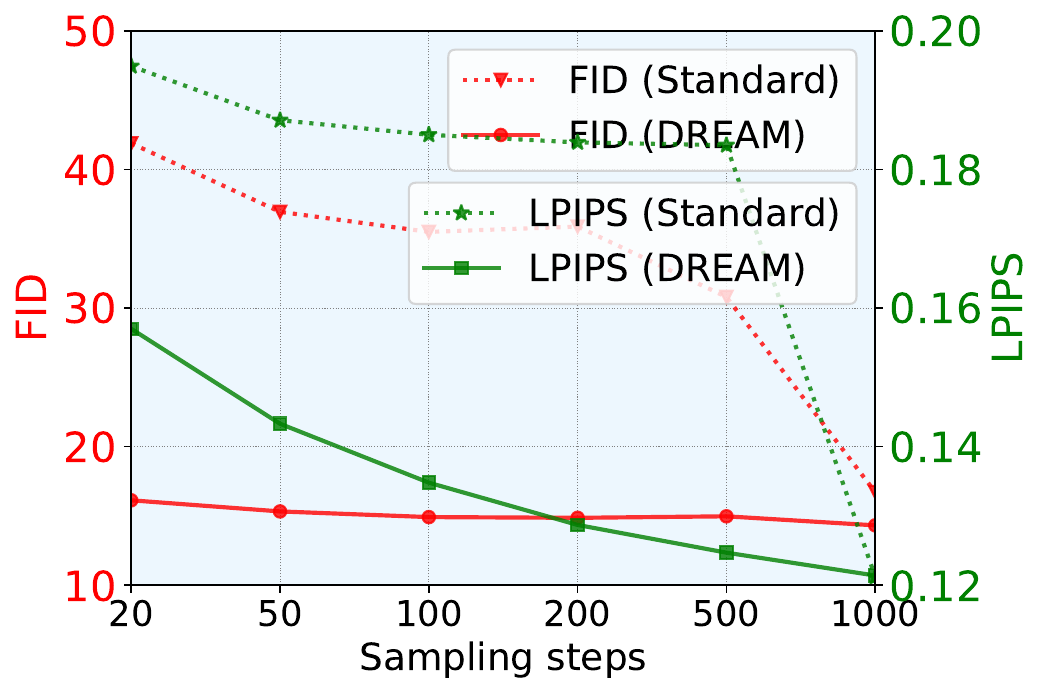}
         \caption{Perception}
         \label{fig:sr3-div-sampling-fl}
     \end{subfigure}
     \vspace{-.28in}
         \caption{Comparison of distortion metrics (left) and perception metrics (right) with varying sampling steps, using  SR3 as a baseline on the DIV2K dataset.}
        \label{fig:sr3-div-sampling}
        \vspace{-.2in}
\end{figure}
\noindent\textbf{Sampling acceleration.} Furthermore, DREAM significantly enhances the efficiency of the sampling process, surpassing the performance of standard diffusion training with a reduced number of sampling steps. \Cref{fig:idm-face-sampling} showcases the capabilities of DREAM using the IDM model on the CelebA-HQ dataset. It compares super-resolution images generated with different numbers of sampling steps, evaluating them against both distortion and perception metrics. While the conventional baseline necessitates up to 2000 sampling steps, DREAM attains superior distortion metrics (an SSIM of 0.73 compared to 0.71) and comparable perceptual quality (an LPIPS of 0.179 versus 0.172) with merely 100 steps, leading to an impressive $20\times$ increase in sampling efficiency. In a similar vein, \Cref{fig:sr3-div-sampling-ps} illustrates the impact of DREAM using the SR3 model on the DIV2K dataset. Here, the images produced with varying sampling steps are again evaluated using both sets of metrics. Standard baselines typically require 1000 sampling steps, but with DREAM, improved distortion metrics (an SSIM of 0.79 versus 0.76) and similar perceptual quality (an LPIPS of 0.127 versus 0.121) are achieved with just 100 steps, resulting in a substantial $10\times$ sampling speedup.

\subsection{Visualization}\label{sec:app-vis}
\begin{figure}[t]
    \centering
    \includegraphics[width=0.99\columnwidth]{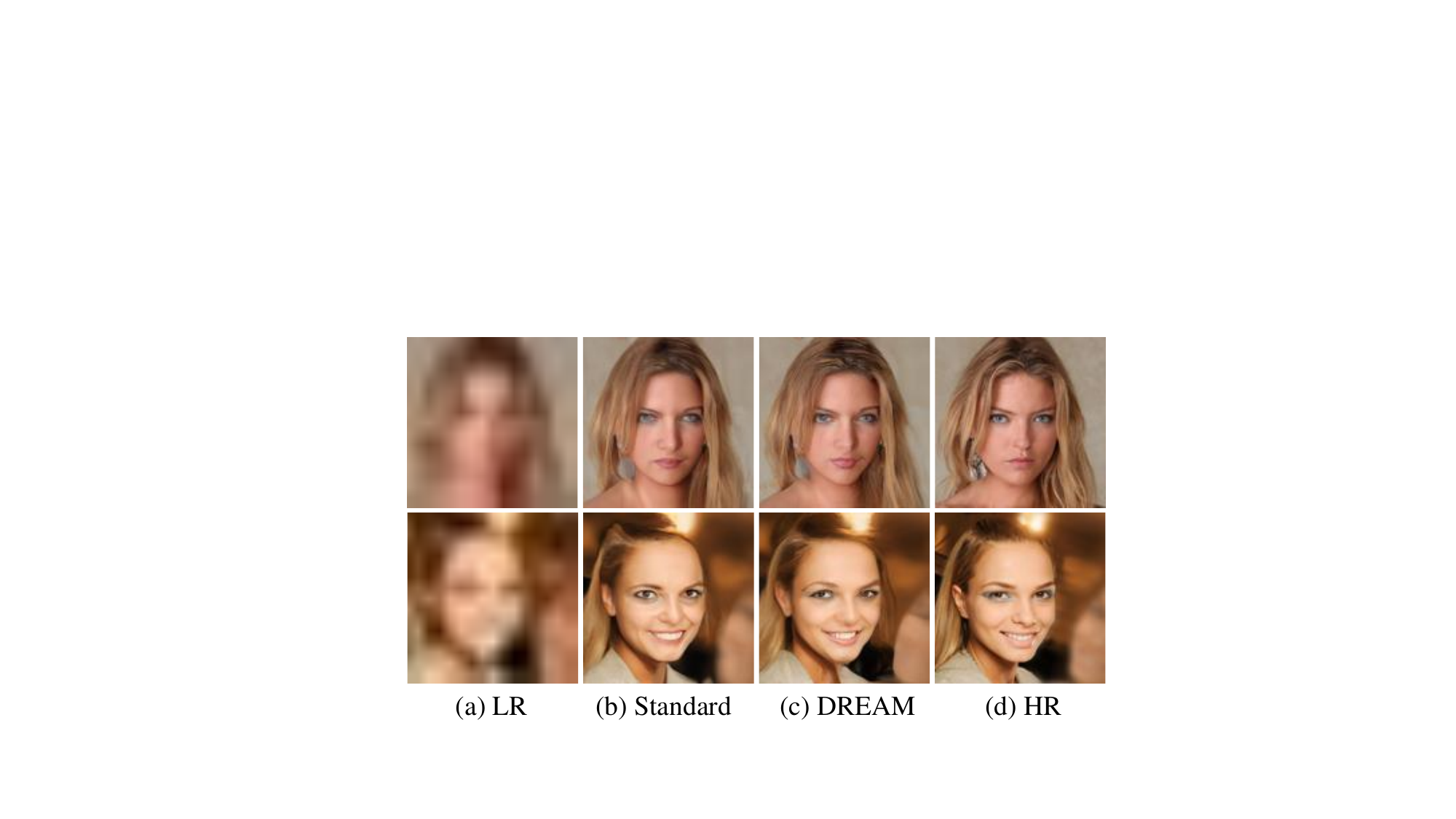}
    \vspace{-.1in}
    \caption{
    Qualitative comparison for $8\times$ SR using SR3~\cite{saharia2022image} on the CelebA-HQ dataset~\cite{karras2017progressive}. Results highlight DREAM's superior fidelity and enhanced identity preservation, leading to more realistic details, such as eye and teeth.} 
        \label{fig:app-face-sr3}
    \vspace{-.1in}
\end{figure}

\begin{figure}[t]
    \centering
    \includegraphics[width=0.99\columnwidth]{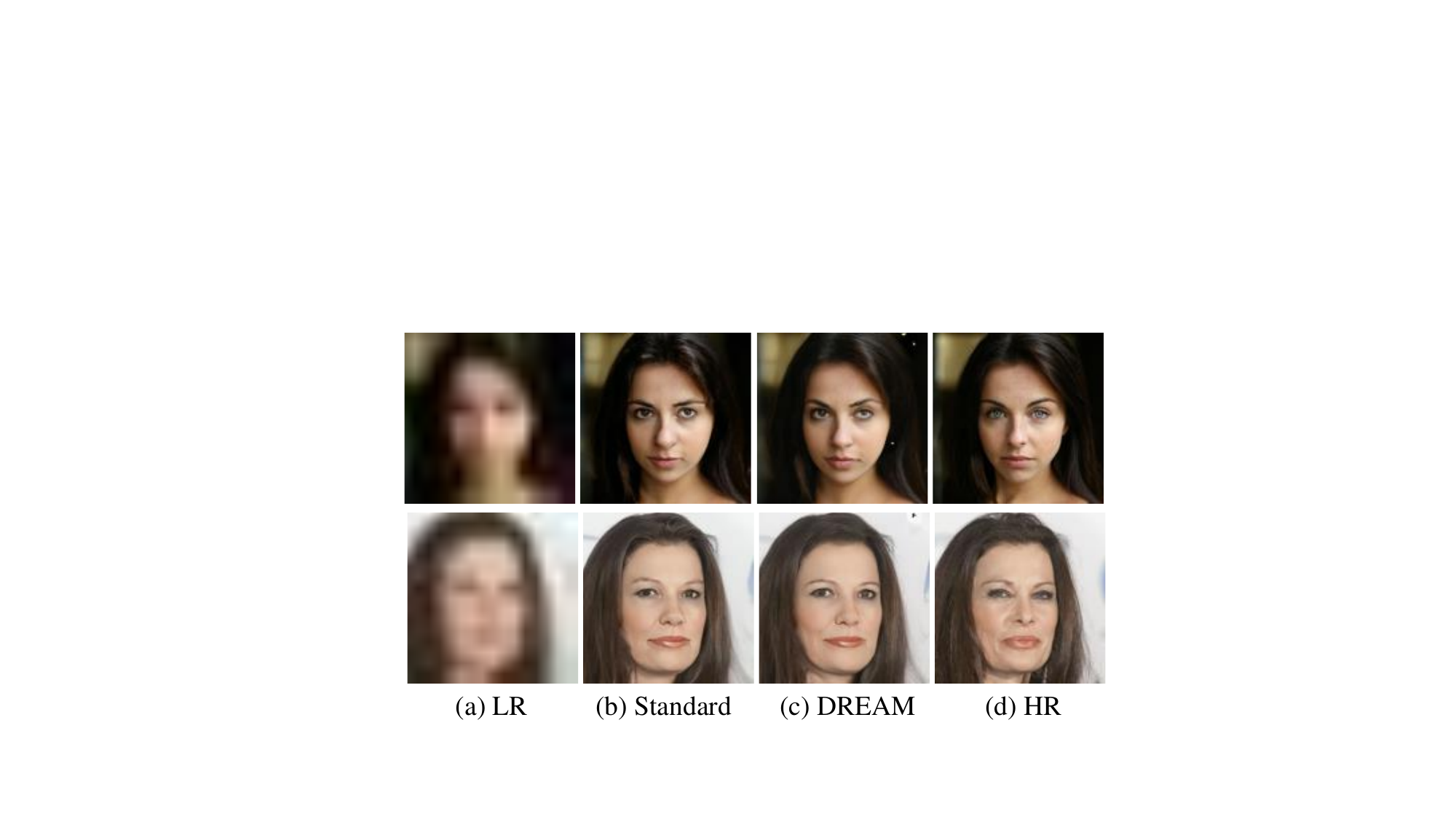}
    \vspace{-.1in}
    \caption{
    Qualitative comparison for $8\times$ SR using IDM~\cite{gao2023implicit} on the CelebA-HQ dataset~\cite{karras2017progressive}. Results highlight DREAM's superior fidelity and enhanced identity preservation, leading to more realistic detail generation in features like nose, and wrinkles.} 
        \label{fig:app-face-idm}
    \vspace{-.2in}
\end{figure}

\paragraph{Face dataset.} In \Cref{fig:app-face-sr3} and  \Cref{fig:app-face-idm}, we provide more representative examples from CelebA-HQ~\cite{karras2017progressive}, employing SR3 and IDM as baselines, respectively. 
% These results again
% validate the remarkable ability of DREAM strategy in synthesizing high-fidelity face images. 

\begin{figure*}[t]
    \centering
    \includegraphics[width=2.08\columnwidth]{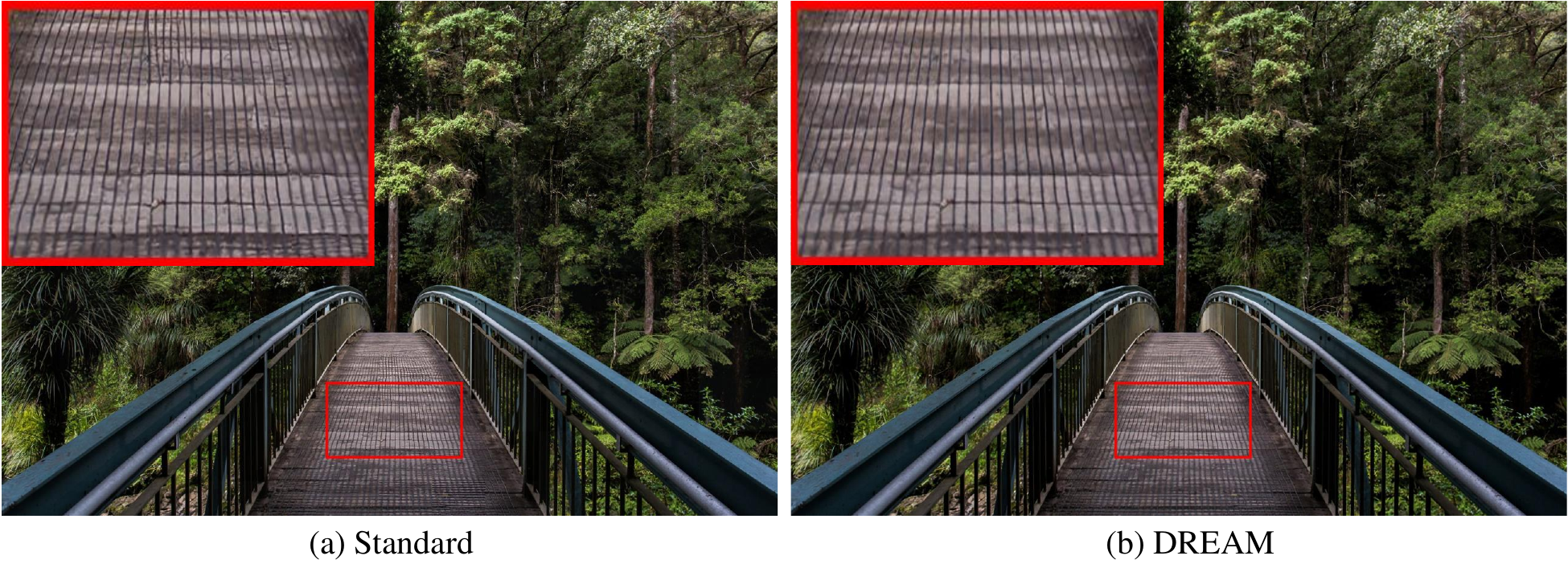}
    \vspace{-.1in}
    \caption{
    Qualitative comparison for $4\times$ SR on DIV2K~\cite{agustsson2017ntire} using SR3~\cite{saharia2022image} model as baseline. \textbf{Left Image:} standard training; \textbf{Right Image:} DREAM training. The model trained under DREAM framework exhibits enhanced fine-grained details and rendering more realistic results, as indicated by the magnified section of the synthesized SR images.}
        \label{fig:sr3-div-exp1}
    \vspace{-.1in}
\end{figure*}

\begin{figure*}[htb]
    \centering
    \includegraphics[width=2.08\columnwidth]{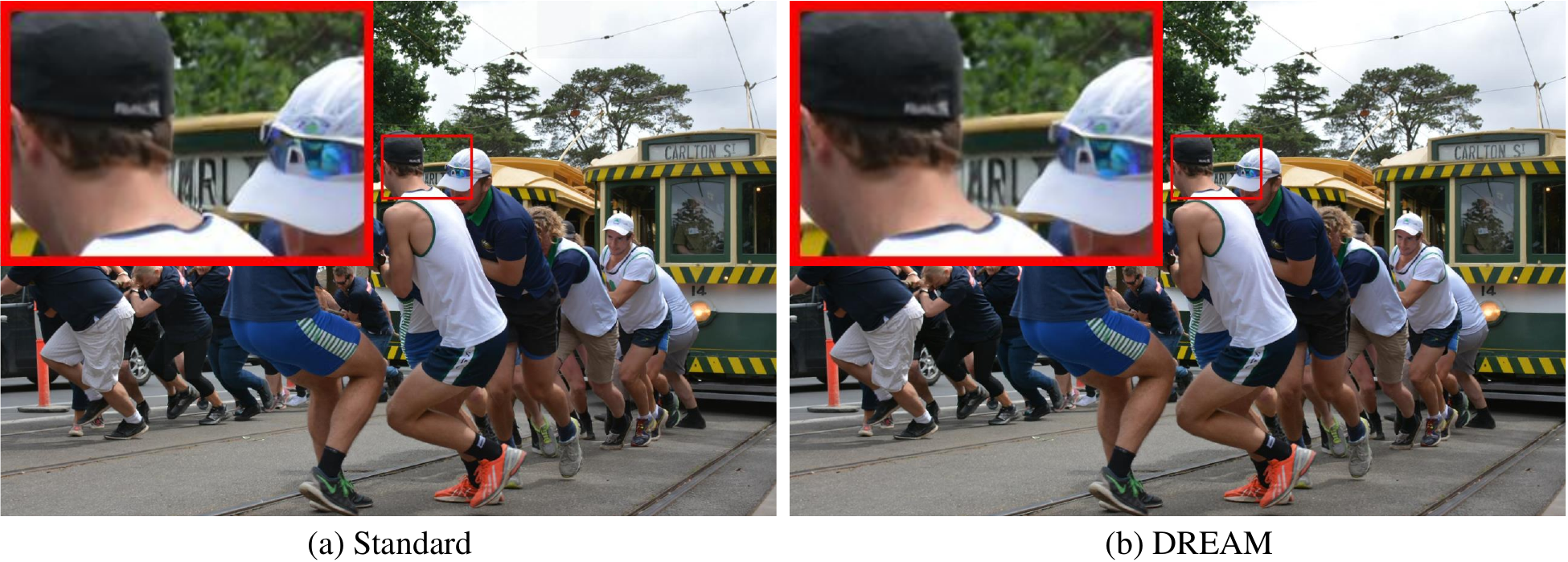}
    \vspace{-.1in}
    \caption{
    Qualitative comparison for $4\times$ SR on DIV2K~\cite{agustsson2017ntire} using SR3~\cite{saharia2022image} model as baseline. \textbf{Left Image:} standard training; \textbf{Right Image:} DREAM training. The model trained under DREAM framework exhibits enhanced fine-grained details and rendering more realistic results, as indicated by the magnified section of the synthesized SR images.}
        \label{fig:sr3-div-exp2}
    \vspace{-.1in}
\end{figure*}

\begin{figure*}[]
    \centering
    \includegraphics[width=2.08\columnwidth]{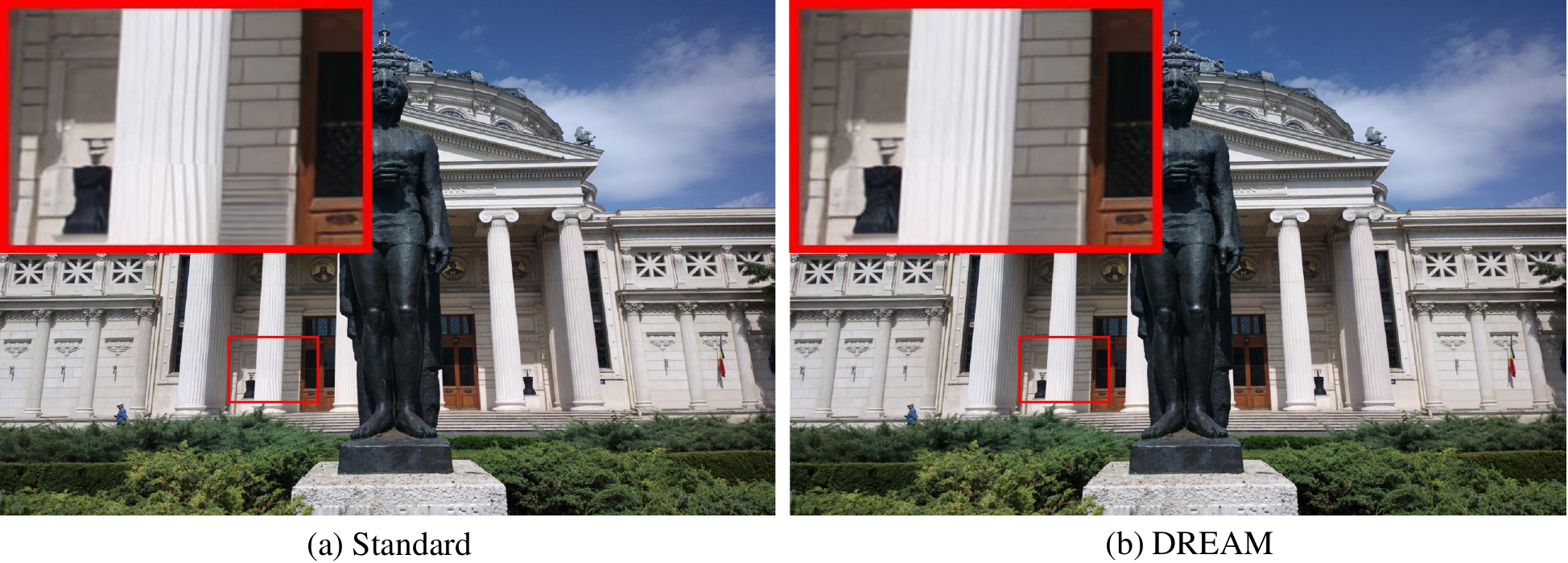}
    \vspace{-.1in}
    \caption{
    Qualitative comparison for $4\times$ SR on DIV2K~\cite{agustsson2017ntire} using SR3~\cite{saharia2022image} model as baseline. \textbf{Left Image:} standard training; \textbf{Right Image:} DREAM training. The model trained under DREAM framework exhibits enhanced fine-grained details and rendering more realistic results, as indicated by the magnified section of the synthesized SR images.}
        \label{fig:sr3-div-exp3}
    \vspace{-.1in}
\end{figure*}

\begin{figure*}[t]
    \centering
    \includegraphics[width=2.08\columnwidth]{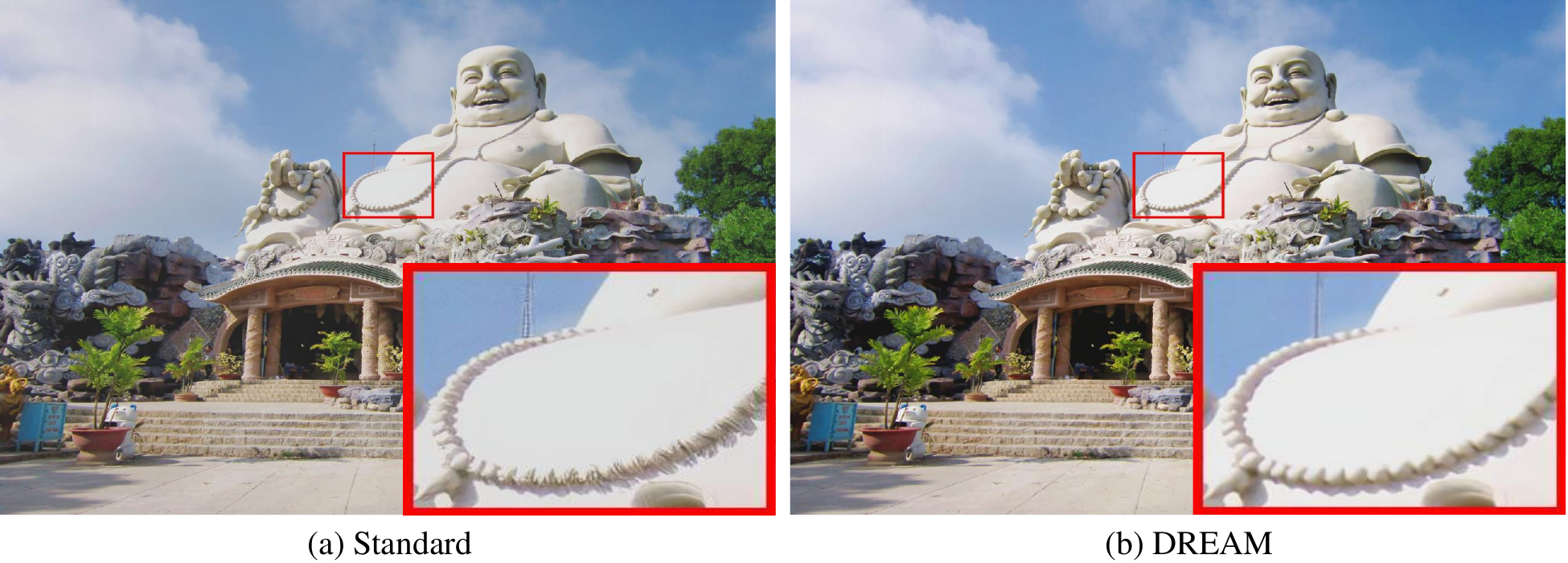}
    \vspace{-.1in}
    \caption{
    Qualitative comparison for $4\times$ SR on DIV2K~\cite{agustsson2017ntire} using ResShift~\cite{yue2023resshift} model as baseline. \textbf{Left Image:} standard training; \textbf{Right Image:} DREAM training. The model trained under DREAM framework exhibits enhanced fine-grained details and rendering more realistic results, as indicated by the magnified section of the synthesized SR images.}
        \label{fig:res-div-exp1}
    \vspace{-.1in}
\end{figure*}

\begin{figure*}[htb]
    \centering
    \includegraphics[width=2.08\columnwidth]{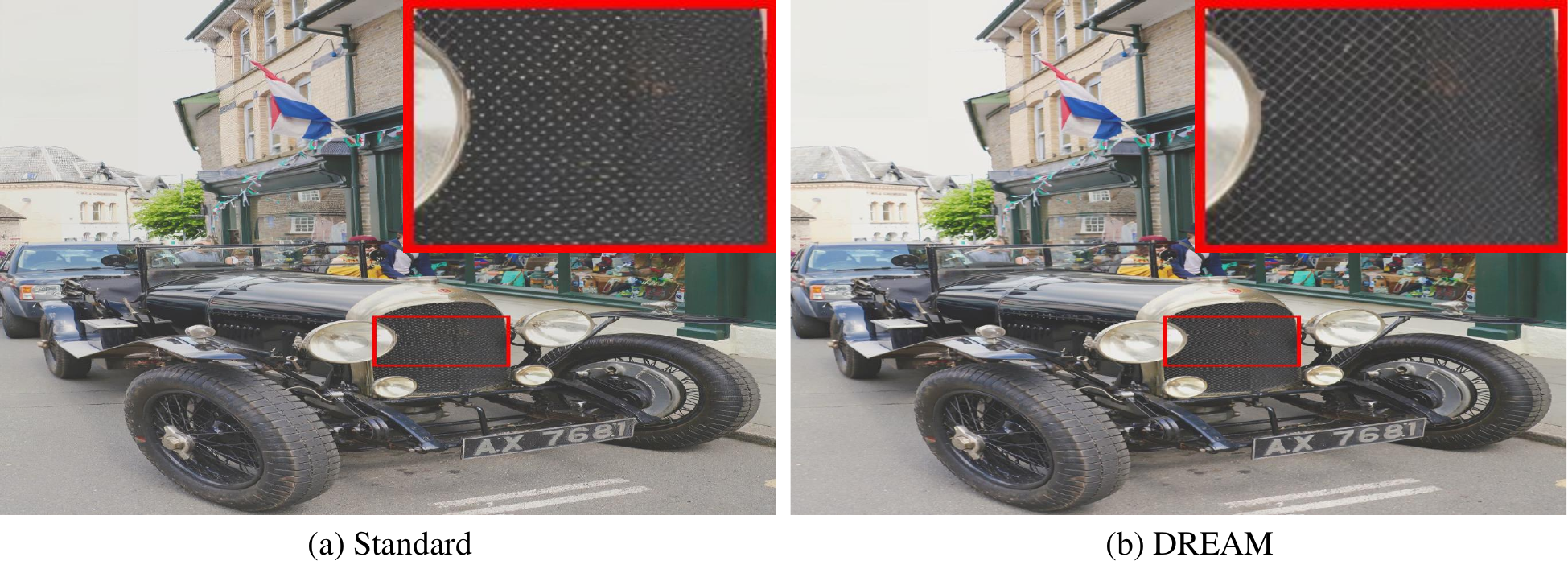}
    \vspace{-.1in}
    \caption{
    Qualitative comparison for $4\times$ SR on DIV2K~\cite{agustsson2017ntire} using ResShift~\cite{yue2023resshift} model as baseline. \textbf{Left Image:} standard training; \textbf{Right Image:} DREAM training. The model trained under DREAM framework exhibits enhanced fine-grained details and rendering more realistic results, as indicated by the magnified section of the synthesized SR images.}
        \label{fig:res-div-exp2}
    \vspace{-.1in}
\end{figure*}

\begin{figure*}[]
    \centering
    \includegraphics[width=2.08\columnwidth]{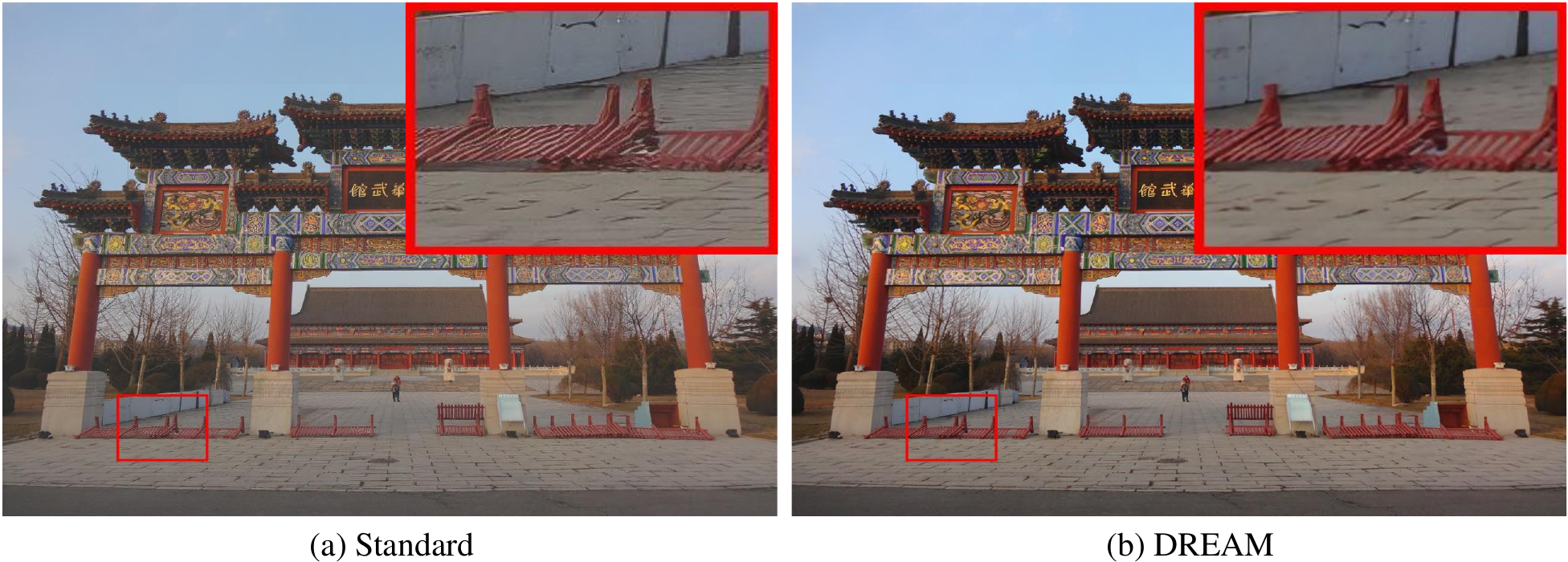}
    \vspace{-.1in}
    \caption{
    Qualitative comparison for $4\times$ SR on DIV2K~\cite{agustsson2017ntire} using ResShift~\cite{yue2023resshift} model as baseline. \textbf{Left Image:} standard training; \textbf{Right Image:} DREAM training. The model trained under DREAM framework exhibits enhanced fine-grained details and rendering more realistic results, as indicated by the magnified section of the synthesized SR images.}
        \label{fig:res-div-exp3}
    \vspace{-.1in}
\end{figure*}

\noindent\textbf{General scene dataset.} To further illustrate the effectiveness of our DREAM  strategy, we present selected examples from the DIV2K~\cite{agustsson2017ntire}. These examples showcase complex image elements such as intricate textures, repeated symbols, and distinct objects. We conduct a comparative visualization of our DREAM strategy against standard training practices, employing the SR3 model as a baseline in \Cref{fig:sr3-div-exp1}, \Cref{fig:sr3-div-exp2} and \Cref{fig:sr3-div-exp3}. Similarly, we use the ResShift model as a baseline in \Cref{fig:res-div-exp1}, \Cref{fig:res-div-exp2} and \Cref{fig:res-div-exp3}. 

All these comparisons unequivocally demonstrate the superior performance of our DREAM strategy.  
\section{Ethic impact}
\label{sec:app-ethic}
This research is applicable to the task of enhancing human facial resolution, a frequent requirement in mobile photography. It does not inherently contribute to negative social consequences. However, given personal security concerns, it is crucial to safeguard against its potential misconduction.

\end{document}